\ifdefined\DRAFTMODE	
\documentclass[preprint,review,3p]{elsarticle}	
\usepackage[sfdefault,light]{FiraSans}	
\usepackage[T1]{fontenc}	

\else	
\documentclass[preprint,review,3p]{elsarticle}
\fi

\usepackage{colortbl}
\usepackage{amsmath}
\usepackage{url}
\usepackage{xspace}
\usepackage{amssymb}
\usepackage{amsfonts}
\usepackage{multirow}
\usepackage{graphicx}
\usepackage{tablefootnote}
\usepackage{enumitem}
\usepackage{balance}
\usepackage{lipsum}
\usepackage{makecell}
\usepackage{textcomp}
\usepackage{flushend}
\usepackage{lipsum}

\usepackage{color}\usepackage{xcolor}\usepackage[normalem]{ulem}

\usepackage{dblfloatfix}
\usepackage{pgfplots,tikz}
\usepackage{booktabs} 

\usepackage{float}

\usepackage{hyperref}
\usetikzlibrary{positioning}
\usetikzlibrary{calc}
\usetikzlibrary{external}
\usepackage{algorithmicx}
\usepackage[noend]{algpseudocode}
\usepackage{subcaption}
\usepackage{mathtools}
\usepackage{bbm}

\DeclareMathOperator*{\argmin}{argmin}
\newcommand{\bst}{\boldsymbol{\theta}}
\newcommand{\bsw}{\boldsymbol{\omega}}
\newcommand{\bsz}{\boldsymbol{z}}
\newcommand{\bsZ}{\boldsymbol{Z}}
\newcommand{\bsa}{\boldsymbol{a}}
\newcommand{\bsp}{\boldsymbol{\pi}}
\newcommand{\bsf}{\boldsymbol{\phi}}
\newcommand{\bss}{\boldsymbol{s}}
\newcommand{\bs}[1]{\boldsymbol{#1}}

\biboptions{longnamesfirst,square,sort,comma}

\usepackage[show]{chato-notes}

\begin{document}
\begin{frontmatter}

    \title{Dynamic Hard Pruning of Neural Networks at the Edge of the Internet}
    \author[IIT]{Lorenzo Valerio\corref{cor1}}\ead{lorenzo.valerio@iit.cnr.it}
    \author[ISTI]{Franco Maria Nardini}\ead{francomaria.nardini@isti.cnr.it}
    \author[IIT]{Andrea Passarella}\ead{andrea.passarella@iit.cnr.it}
    \author[ISTI]{Raffaele Perego}\ead{raffaele.perego@isti.cnr.it}
    \address[ISTI]{ISTI-CNR, via Moruzzi 1, 56124 Pisa, Italy}
    \address[IIT]{IIT-CNR, via Moruzzi 1, 56124 Pisa, Italy}
    \cortext[cor1]{Corresponding author}


\begin{abstract}
Neural Networks (NN), although successfully applied to several Artificial Intelligence tasks, are often unnecessarily over-parametrised. In edge/fog computing,  this might make their training prohibitive on resource-constrained devices, contrasting with the current trend of decentralising intelligence from remote data centres to local constrained devices. Therefore, we investigate the problem of training effective NN models on constrained devices having a fixed, potentially small, memory budget. We target techniques that are both resource-efficient and performance effective while enabling significant network compression. Our Dynamic Hard Pruning (DynHP) technique incrementally prunes the network during training, identifying neurons that marginally contribute to the model accuracy.  DynHP enables a tunable size reduction of the final neural network and reduces the NN memory occupancy during training. Freed memory is reused by a \emph{dynamic batch sizing} approach to counterbalance the accuracy degradation caused by the hard pruning strategy, improving its convergence and effectiveness. We assess the performance of DynHP through reproducible experiments on three public datasets, comparing them against reference competitors. Results show that DynHP compresses a NN up to $10$ times without significant performance drops (up to $3.5\%$ additional error w.r.t. the competitors), reducing up to $80\%$ the training memory occupancy.
\end{abstract}

\end{frontmatter}


\section{Introduction}
\label{sec:intro}

In the last years, AI solutions have been successfully adopted in a variety of different tasks. Neural networks (NN) are among the most successful technologies that achieve state-of-the-art performance in several application fields, including image recognition, computer vision, natural language processing, and speech recognition. The main ingredients of NNs success are the increased availability of huge training datasets and the possibility of scaling their models to millions of parameters while allowing for a tractable optimization with mini-batch stochastic gradient descent (SGD), graphical processing units (GPUs) and parallel computing. Nevertheless, NNs are characterized by several drawbacks, and many research challenges are still open. Recently, it has been proven that NNs may suffer over parametrization \cite{Han2015,ullrich2017soft,molchanov2017variational} so that they can be pruned significantly without any loss of accuracy. Moreover, they can easily over-fit and even memorize random patterns in the data \cite{zhang2016understanding} if not properly regularized.

NN solutions are typically designed having in mind large data centres with plenty of storage, computation and energy resources, where data are collected for training, and input data are also collected at inference time. This scenario might not fit emerging application areas enabled by the widespread diffusion of IoT devices, commonly referred to as \emph{fog computing environments}. Typical application areas are smart cities, autonomous vehicular networks, Industry 4.0, to name a few. IoT devices in fog environments generate huge amounts of data that, for several reasons, might be impossible or impractical to move to remote data centres both for training and for inference. Typically, real-time or privacy/ownership constraints on data make such an approach unfeasible. Therefore, knowledge extraction needs to leverage distributed data collection and computing paradigms, whereby NNs are ``used" in locations closer to where data are generated, such as fog gateways or even individual devices such as tablets or Raspberry PIs. Unfortunately, with respect to data centres, these devices have much more limited computational power, memory, network, and energy capabilities. In these contexts, the exploitation of models trained ``offline'' asks for a significant amount of memory -- from hundreds of MBytes to GBytes -- and their use, i.e., inference, requires GFLOPs of computation. As an example, the inference step using the AlexNet network \cite{krizhevsky2012imagenet} costs 729 MFLOPs (FLoating-point OPerations), and the model requires $\sim250MBytes$ of memory to be stored.
Furthermore, resource-constrained devices should deal with limited energy availability (e.g., some devices might be battery powered). Interestingly, the energy consumption of such devices is dominated by memory access - one DRAM access costs two orders of magnitude more than one SRAM access, and three orders of magnitude more than one CMOS add operation \cite{Modarressi:2018aa}. These limitations jeopardize the exploitation of large models trained ``offline'' in resource-constrained devices characterizing the edge/fog computing paradigms. Training DNNs on such devices is even more challenging, as the training phase is typically much more resource-hungry than the inference phase.

Nowadays, several researchers are investigating the use of neural networks on resource-constrained devices. The effort is focused on enabling their use at both training and inference time on this kind of devices by limiting -- or possibly totally avoiding -- the loss in performance introduced by the reduced memory and computational capacity of the devices targeted. These approaches can be broadly classified into two research lines. On the one hand, we have methods that, given an already trained neural network, try to reduce its size or distribute it on several devices collaborating during the inference phase. Three main lines have been investigated under this main approach: pruning, quantization, and knowledge distillation \cite{Guo2016b,Han2016a,Lin2016,Hinton2015}. On the other hand, there are a few proposals that work at training time. These methods employ neural compression techniques \cite{Srinivas2017,Louizos2017} or neural architecture search \cite{JMLR:v20:18-598} to identify effective configurations that actively reduce the size of the model. It is important to note that techniques following the latter approach typically measure the compression efficacy as the average number of neurons simultaneously active during the training.\footnote{Here we refer to neurons for simplicity of presentation, but the same argument can be extended to other groups of parameters, such as filters, channels, etc., according to the kind of network and pruning approach at hand.} Neural compressors working at training time during an epoch switch off some neurons according to a given criterion. However, we have \emph{soft} pruning techniques that allow neurons to be switched on again after they have been switched off. Conversely, this does not happen when \emph{hard} pruning techniques are adopted: neurons that are switched off are lost. While soft pruning guarantees more flexibility during training, it does not reduce the memory occupation of the model during the training process, as information about switched-off neurons has to be always kept from epoch to epoch.

This paper aims to design new hard pruning techniques for learning compressed neural networks on resource-constrained devices typical of edge/fog environments. We claim that learning compressed neural networks directly on edge/fog devices is of paramount importance to achieve pervasive -- both effective and efficient -- neural network-based AI solutions in these environments. For this reason, we want to target techniques that are both resource-efficient and allow comparable levels of performance with respect to conventional neural network training algorithms while enabling significant levels of compression of the network. We target the goal by assuming that the training of a neural network in edge/fog devices relies on a fixed -- and often small -- budget of memory that can be used to perform the process. Our assumption comes from observing that the edge/fog paradigm is characterized by moving the computation close to the data source~\cite{Garcia-Lopez:2015aa,Conti:2017aa}. In this scenario, the devices employed can perform many other operations in parallel w.r.t. training a neural network, e.g., operations related to data gathering/indexing/storing~\cite{Barbalace:2020}. The edge/fog scenario is different from a standard cloud environment, where we can assume to have servers fully available to train the neural network. We thus propose a new technique based on an effective compression of the network during the training process. Our novel technique, called \emph{Dynamic Hard Pruning} (DynHP), prunes incrementally but permanently the network by identifying neurons that contribute only marginally to the model accuracy. By doing so, DynHP enables a significant and controllable reduction of the size of the final neural network learned. Moreover, by hard pruning the neural network, we progressively reduce the memory occupied by the model as the training process progresses. Finally, our solution is designed to train and prune the NN under a fixed memory budget, which might be an important feature if we consider a scenario where an edge/fog device, e.g., Nvidia Xavier or Jatson Nano, has to run multiple, e.g., containerized, training processes of different NNs on local data.

However, hard pruning neurons also brings a side effect related to the convergence of the training process to accurate solutions. Precisely, it slows down the convergence and makes stochastic gradient descent more susceptible to get stuck in poor local minima. We capitalize on the increasing amount of memory saved during training to introduce a dynamic sizing of the mini-batches used to train the network to avoid this limitation. Our dynamic batch sizing technique enables direct control of the convergence and the final effectiveness of the network by varying the amount of data seen in a batch. DynHP tunes the size of the batch dynamically,  epoch by epoch, by computing its optimal size as a function of the variance of the gradients observed during the training. Moreover, we dynamically adjust the amount of data seen by considering the constraint on the total memory available for the training process.  Our proposal thus effectively reuses the memory progressively saved with hard pruning to increase the size of batches used to estimate the gradient and improves convergence speed and quality. 

The result achieved by DynHP is three-fold. First, the training of compressed neural networks can be done directly on resource-constrained edge/fog devices. Second, our technique minimizes the performance loss due to network compression by reusing the memory saved for the network to increase the batch size and improve convergence and accuracy. Third, we explicitly target the training of neural networks under hard memory constraints by dynamically optimizing the learning process on the basis of the amount of memory available. 

We assess our DynHP on three public datasets, MNIST, Fashion-MNIST and CIFAR-10, against state-of-the-art competitors. Our reproducible experiments show that DynHP can effectively compress from $2.5$ to $10$ fold a DNN  without any significant effectiveness degradation, i.e., $3.5\%$ of additional misclassification error w.r.t. competitors. Moreover, beyond obtaining highly accurate compressed models, our solutions dramatically reduce the overall memory occupation during the entire training process, i.e., we use from $\sim 50\%$ to $\sim 80\%$ less memory to complete the training.

The paper is structured as follows: Section \ref{sec:related} discusses related work, while Section \ref{sec:background} presents the background of our work. Moreover, Section \ref{sec:HP} introduces the hard pruning of neural network, while Section \ref{sec:dynhp} presents the dynamic hard pruning technique. Section \ref{sec:experiments} presents and discusses the experimental results on three public datasets. Finally, Section \ref{sec:conclusion} concludes the article and draws some future work.


\section{Related Work} 
\label{sec:related}

It is a known fact that neural networks models are often over-parametrised, i.e., the number of the model's parameters that must be trained and the resulting complexity of the model exceed the one needed by the problem at hand. Due to this fact,  several methods have been proposed to reduce the size of a neural network model (i.e., lower the number of parameters). The challenge of all such methods is finding a way to obtain a final network model with only the necessary number of parameters needed to achieve the same accuracy as the ``over-parametrised model''. In the literature,  three main directions have emerged, along with we can categorise the algorithmic solutions that cope with such a problem. 

The first category includes solutions that perform the compression \emph{after} the training phase. Once the over-parametrised model is trained, these approaches exploit and/or manipulate it to obtain a smaller one with the same generalisation capabilities. One of the seminal works of this category is the so-called \emph{Knowledge Distillation method}~\cite{Hinton2015,Ba2013} that uses an already trained massive and complex Deep Neural Network (called teacher network) to guide the training of a smaller one (called student network) whose task is to learn to mimic the behaviour of the teacher network. Other approaches propose to prune useless or redundant connections of a Deep Network. Han \emph{et al.} propose a pruning methodology in three steps: first, they train the network, then they prune redundant weights, and finally, they re-train the pruned network  \cite{Han2015}.
Similarly, Guo \emph{et al.} propose a compression method made of two phases: pruning and splicing~\cite{Guo2016b}. Briefly, in the pruning phase, the connections deemed as useless are removed. However, to recover from a possibly aggressive pruning that causes a sheer drop in the accuracy performance of the model, the authors propose a way to reactivate some of the connections that might be necessary to the generalisation capabilities of the network. Such an idea has also been explored in~\cite{Han2016a,Jin2016}. Despite the very impressive results both in terms of compression and accuracy, all these techniques require several training phases through which the compression is incrementally done. Therefore, in terms of memory usage, the advantages become significant only when the final model is deployed and used, while the training phase is still quite resource-demanding. 

Alternatively, instead of pruning the connections of an already over-parametrised model, other approaches propose acting on the weights' numerical representation. Note that these solutions can be applied to the network either \emph{before} or \emph{after} the training phase. Han \emph{et al.} propose a three-stage pipeline - \emph{pruning, trained quantisation and Huffman coding} - that work together to reduce the storage requirement of neural networks by 35x to 49x without affecting their accuracy \cite{Han2015}. Hubara \emph{et al.} propose a method to train a Deep Network with binary weights and activation functions \cite{Hubara2016}. They can run the network 7x faster, achieving \emph{nearly} state of the art accuracy on relatively small datasets like MNIST and CIFAR-10. However, with more challenging datasets, like ImageNet, this approach does not obtain significant accuracies. Another work that follows the same principles proposes to use a fixed point quantisation for representing the weights~\cite{Lin2016}. The main drawback is that it can be applied only to an already trained model. Similarly, \cite{Courbariaux2015} propose to use binary weights and a custom forward and backward pass through which they reach nearly state of the art accuracies on relative simple datasets (e.g. MNIST, CIFAR-10). To summarising, most of these quantisation/sparsification solutions are orthogonal to the one proposed in this paper because they can be applied, at least in principle, to our method for further reducing the memory occupation of the model at hand.

Finally, the third category includes solutions that compress the model \emph{during} the training phase. According to the literature, the compression at training time is mainly based on weights pruning, which can be accomplished in two different ways. On the one hand, some works prune single weights, sparsifying the model. On the other hand, some solutions prune neurons by removing an entire row of the weights matrix. In the first line of research, Bellec \emph{et al.} train directly a sparsely connected neural network and replace weights that reach zero with new random connections \cite{Bellec2017}. They experimentally show that the technique can be effective in learning sparse networks with a negligible loss in performance. In the same line, Narang \emph{et al.} integrate magnitude-based pruning into training showing that the model size can be reduced by 90\% with a speed-up from $2$x to $7$x \cite{Narang2017}. Srinivas \emph{et al.} and Louizos \emph{et al.} learn gating variables with the aim at minimising the number of nonzero parameters of the network \cite{Srinivas2017,Louizos2017}. Recently, Lin \emph{et al.} proposed a model compression method that generates a sparse trained model without additional overhead \cite{DBLP:conf/iclr/LinSBDJ20}. The proposed technique works i) by allowing a dynamic allocation of the sparsity pattern and ii) by incorporating feedback signals to reactivate prematurely pruned weights. Results on CIFAR-10 show the effectiveness of the technique in learning effective sparse networks.

In the second line of research, Srinivas \emph{et al.} rely on the fact that similar neurons are redundant. They compute the similarity of each pair of neurons and remove, one neuron at a time, the most similar ones.~\cite{Srinivas2015}. Another structured approach is the one by Molchanov \emph{et al.}, where authors propose a structured dropout technique to prune convolutional kernels in neural networks to make them more efficient \cite{Molchanov2016}. On the same line, Neklyudov \emph{et al.} propose a structured Bayesian dropout that allows to sparsify the network at training time \cite{Neklyudov2017}. Finally, Frankle \emph{et al.} \cite{Frankle2018} introduce the concept of \emph{winning tickets}, discussing the existence of one or more small networks inside a bigger one that is sufficient to reach good accuracy with fewer parameters. The authors propose a method to discover such subnetworks. All these methods have the advantage to produce a compressed and accurate network at the end of the training phase.

In the techniques above, only the work by Louizos \emph{et al.} allows reducing the memory size of a network at training time \cite{Louizos2017}. In fact, pruning the weights in a scattered way - as most methods do - is not helpful for memory-saving purposes at training time because of the way matrices are represented in memory. Conversely, pruning the neurons is much more effective since we can actually reduce the size of the weight matrix. In this paper, we build on top of the gating mechanism proposed by Louizos \emph{et al.} \cite{Louizos2017}. Specifically, our novel technique, called Dynamic Hard Pruning (DynHP), prunes incrementally but permanently the network by identifying neurons that contribute only marginally to the model accuracy. By doing so, DynHP enables a significant reduction of the size of the final neural network learned. Moreover, by hard pruning the neural network, we significantly reduce the memory occupied by the model during the training process. The available memory is thus reused to minimise the accuracy degradation caused by the hard pruning strategy by adaptively adjusting the size of the data provided to the neural network to improve its convergence and final effectiveness.


\section{Background}
\label{sec:background}
In this section we first introduce the notation used in the rest of the paper, then we discuss the state of the art techniques for soft pruning neural networks (during training) that inspired our solution. For the sake of clarity, we report the only details that are necessary to make this paper self-contained. 

\subsection{Preliminaries}
We assume to have a dataset $\mathcal{D}=\{(x_i,y_i)\}_{i=1}^N \in \mathbb{R}^{N\times d+k}$ containing  $N$ i.i.d. $d$-dimensional observations $x_i \in \mathcal{X}\subseteq\mathbb{R}^d$, each one accompanied by a label $y_i \in \mathcal{Y}\subseteq\mathbb{R}^k$. Note that  we target supervised learning  problems with one or more labels per observation. The neural network model with weights $\bsw$ is denoted by the function $h:\mathbb{R}^{d}\rightarrow \mathbb{R}^k$. 
Additionally, in the following, we refer to the set of neurons with the symbol $\bst$. 
Let $\ell:\mathbb{R}^{k} \times \mathbb{R}^{k}\rightarrow\mathbb{R}$ be the loss function  used to evaluate the prediction accuracy of the model $h$. The operator $\odot$ is the Hadamard product, i.e., the element-wise product between vectors or matrices. Finally, let us denote with $\|\cdot\|_p$ the generic $p$-norm. 

\subsection{Soft Pruning of Neural Networks}
\label{ssec:SP}
Our solution is inspired by the technique proposed in \cite{Louizos2017}, from now on referred to as \emph{Soft Pruning} (SP). SP is based on the following \emph{Regularized Empirical Risk minimization problem}:
\begin{eqnarray}
\label{eq:obj}
\mathcal{L}(\bsw) = \frac{1}{N}\sum_{i=1}^N \ell(h(x_i;\bsw),y_i) +\lambda\left\|\bsw \right\|_0
\end{eqnarray}
$$\bsw^* = \argmin_{\bsw} \mathcal{L}(\tilde\bsw)$$
where the first component of $\mathcal{L}(\bsw)$ is the average loss of model $h$ over the dataset, and the second component is the norm $L_0$ acting as a regularizer tuned with the $0<\lambda <1$ parameter. Since the $L_0$-norm counts the number of non-zero parameters of the neural network, using it as a regularization term drives the learning algorithm towards solutions having a small number of connections whose weight is non-zero. We note that as soon as \emph{all} the weights of a neuron become equal to zero, the neuron itself could be pruned and the size of the network reduced. However, the above formulation does not provide any fine-grained control over the turning-off of all the weights belonging to individual neurons, and the resulting pruning strategy is not particularly effective. 
A second drawback is that the $L_0$-norm of the weights is not a differentiable function, which prevents the straightforward usage of any gradient-based optimization method from training the neural network.

To overcome the above problems, Louizos \emph{et al.} propose to approximate the $L_0$-norm with an equivalent and differentiable function \cite{Louizos2017}.
They propose a re-parametrization of $\bsw$ such that each parameter $\omega_j$ is defined as follows: $$\omega_j = \tilde{\omega_j}z_j$$ where $z_j\in\{0,1\}$ is a binary gate that controls the activation of the j-th parameter. Therefore the $L_0$-norm becomes: 
\begin{equation}
\label{eq:l0-first}
\|\bsw\|_0 = \sum_{j=1}^{|\bsw|} z_j
\end{equation}
The main intuition in \cite{Louizos2017} is to model the gates as Bernoulli random variables:
$$q(z_j|\pi_j)=\mathrm{Bern}(\pi_j)$$ 
Precisely, each binary gate $z_j$ has a probability $\pi_j$ of being active.
Adopting a probabilistic representation for the gates means that weights $\bsw$ and the loss value $\ell(,)$ become random variables. 
Therefore, the objective function in Equation~\ref{eq:obj} needs to become an average, as shown in Equation~\ref{eq:obj-avg}:~\footnote{For simplicity we omit most of the technicalities needed to obtain such formulation. The interested reader can find more information in \cite{Louizos2017} and \cite{Kingma2015}.}
\begin{eqnarray}
\mathcal{L}(\tilde{\bsw},\bsp) = \mathbb{E}_{q(\bsz|\bsw)}\left[\frac{1}{N}\sum_{i=1}^N \ell(h(x_i;\tilde{\bsw}\odot \bsz),y_i)\right] +\lambda\sum_{j=1}^{|\bsw|} \pi_j
\label{eq:obj-avg}
\end{eqnarray}

$$\tilde{\bsw}^*, \bsp^* = \argmin_{\tilde{\bsw},\bsp} \mathcal{L}(\tilde{\bsw},\bsp)$$

In this way the learning process is affected by the number of active binary gates and, in particular, we are minimizing at the same time both the generalization error and the sum of the probabilities of the gates. However, in this form the function $\mathcal{L}(\tilde{\bsw},\bsp)$ is not yet suitable for efficient gradient computation due to the fact that the Bernoulli distribution is a discrete function and, consequently, it prevents the smoothness of $\mathcal{L}(\tilde{\bsw},\bsp)$. The solution proposed in \cite{Louizos2017} is to substitute the  Bernoulli distribution used for modelling the gates with the Hard Concrete Distribution \cite{Maddison2017} which is its continuous and differentiable approximation. 
Skipping all the technical steps, the final and differentiable version of  $\mathcal{L}(\tilde{\bsw},\bsp)$, denoted as $\mathcal{R}(\tilde{\bsw},\bsf)$ has the following form:
\begin{equation}
\label{eq:regprob}
\mathcal{R}(\tilde{\bsw},\bsf) = \frac{1}{T}\sum_{l=1}^T \left(\frac{1}{N} \left( \sum_{i=1}^N \ell(h(x_i;\tilde{\bsw}\odot \bsz^{(t)}),y_i)\right) +\lambda\sum_{j=1}^{|\bsw|} \left( 1-Q_{s_j}(0|\phi_j)\right)\right)
\end{equation}
where $Q_{s_j}(0|\phi_j)$ is the CDF of the Hard Concrete distribution.\footnote{Here $s\sim q(\bss|\bsf)$ and $q(\bf{s})$ is the distribution function of $\bf{s}$ with parameter $\phi$. More details in \cite{Louizos2017}.} Equation~\ref{eq:regprob} assumes that training occurs over $T$ epochs, and in each epoch, a random draw for the activation gates is used. The first part of the equation is thus the average over the samples of the average losses obtained at each round, which is a standard estimator for the average value of the average loss. The second part is the average value of the number of non-zero gates. We minimize the objective function with respect to both $\tilde{\bsw}$ and $\bsf$, meaning that we learn, at the same time, the parameters ($\tilde\bsw$) and how many of them are, on average, really useful for the good predictions ($\bsf$). 
\subsection{Discussion}
Summarising, according to SP, each neuron (or a weight) is associated with a gate ($\bs{z}$) that is active (i.e., $\bs{z}>0$) with a probability controlled by the parameter $\phi$. The value of $\phi$ varies during training according to the learning process. Precisely, they are trainable parameters like the weights of the NN, and they get updated through backpropagation. The concrete effect of the gates is to modulate the output of the corresponding neurons (or weights), similarly to what happens with Dropout~\cite{Gal2015} but with the difference that the activation probability is part of the training and a gate can take values in the range $[0,1]$. In this way, during training, the neurons that contribute the most to the generalization capabilities of the NN will have a higher probability of remaining active, i.e., their parameter $\phi$ grows towards 1, while the activation probability of those that are considered ``redundant'' will decrease towards 0.
However, although the use of this technique allows obtaining a compressed version of the initial network maintaining almost the same generalization performance, it is fundamental to point out that, during the training, such compression is just virtual. In fact, according to the gating mechanism, the number of active neurons decreases only on average, meaning that in practice, the neurons switched off in one epoch might return active during the next one and are not removed from the network. 

This characteristic represents a major shortcoming for our purposes of devising a learning solution suitable for training models with limited memory footprints.

\section{Dynamic Hard Pruning of Neural Networks}

In this section, we introduce and discuss our \emph{Dynamic Hard Pruning} technique for training and compressing a neural network at a fixed memory budget. For the sake of clarity, we split the presentation of our solution into two parts. In the following, we first present the \emph{hard pruning} technique that we use to \emph{ablate} parts of the neural network during the training process. We then introduce our mechanism of \emph{dynamic batch-sizing} that we use to drive the overall training process to achieve a significant reduction of the neural network with no loss of accuracy. The main idea is thus to remove less useful neurons at each epoch to save memory. As this may reduce accuracy, we exploit the freed memory within a dynamic batch size to optimize the performance of the pruned network.

\subsection{Hard Pruning}
\label{sec:HP}
Our \emph{Hard Pruning} mechanism (HP) is based on a ``one-way-only strategy'' aimed at identifying and removing the less useful neurons together with all their inward and outward connections. Thanks to HP, the size of the network monotonically decreases as training progresses, and we achieve a corresponding incremental reduction of the network's memory footprint. Differently from SP, the ablation of neurons performed in one epoch cannot be subsequently reverted. However, permanently deleting a part of the network can be highly detrimental to its performance. Thus, to avoid disrupting the learning capabilities of the network, we have to carefully identify the neurons whose removal does not harm, or harms only minimally, the quality of the training process. To this end, we exploit the weights gating mechanism discussed above and collect detailed statistics on gates activation during a fixed observation time window, e.g., a training epoch. At the end of each epoch, we compute the average activation rate $a_j\in[0,1]$ for each gate $j$, and we use such values to identify the least active neurons to be pruned. More formally, let  $\bsZ=\{0,1\}_{j=1}^{|\bst|}$ be a  vector that stores the binary information regarding the status (i.e., active/inactive) of each neuron of a layer $l$.\footnote{For the sake of clarity, the description refers to a single layer but its extension to all the layers of the neural network is straightforward.} Vector $\bsZ$ is updated at the end of each epoch to always record the active neurons in the layer. Moreover, let  $\bsa$ be a vector recording the activation rate for each neuron in the layer during an epoch. $\bsa$ is computed as follows:

\begin{equation}
\bsa = \frac{1}{E}\sum_{e=1}^E \bsz_e
\end{equation}
where $\bsz$ is the number of times the random gates were active during an epoch and $E$ is the total number of gate's state observations (i.e. active or inactive) during a training epoch. To identify the less active neurons in the layer  we use a hard thresholding function $g(\cdot,\gamma)$ with fixed threshold $\gamma$:
\begin{equation}
g(x,\gamma)=\left\{
\begin{array}{lr}
1 & \mathrm{if}\ x \geq \gamma \\
0 & \mathrm{otherwise}
\end{array}\right.
\label{eq:gamma}
\end{equation}
By applying element-wise  function $g(\cdot,\gamma)$ to  vector $\bsa$ we obtain a binary vector $$\hat{\bsz}=g(\bsa,\gamma)$$ that identifies the most active neurons in the layer. At the end of the epoch we use this information to update  the status of the neurons stored in $\bsZ$:
$$\bsZ = \bsZ \odot \hat{\bsz}$$
In addition, vector $\bsZ$ is used to create a binary matrix $M$, $$M=\bsZ\cdot\mathbbm{1}^T$$ where $\mathbbm{1}$ is a vector with all elements equal to $1$ and the same dimension of the successive layer $l+1$. 
 $M$ is finally used to set to zero all the weights corresponding to the deactivated neurons. 
This HP step is repeated at the end of each epoch for all the layers of the network.

\subsection{Dynamic Hard Pruning}
\label{sec:dynhp}
We now present our \emph{Dynamic Hard Pruning} technique.
As one might expect, our HP technique results in an effective reduction of the size of the neural network as training progresses, but it has a significant impact on the convergence of the training process, possibly leading to a worse accuracy with respect to the SP technique, as we will show in Section~\ref{sub:RQ1_results}.

The worse performance is mostly due to the interplay between the hard pruning and the training processes: while the SGD algorithm tries to reduce error, the HP step performed at each epoch may have the opposite effect of increasing the error due to the ablation of parts of the network. Moreover, an additional difficulty regards the relative speed of these two contrasting processes: if pruning is too fast due to an aggressive setting of the parameters involved, the network loses most of its expressiveness with the consequence of converging to a poor model. Conversely, if the pruning is not incisive enough, the resulting model is still unnecessarily over-parametrized. Therefore, it is crucial to find a trade-off between these two different dimensions involved.

Several hyper-parameters drive the dynamics of the pruning and the learning process, thus actively controlling the trade-off above. On the one hand, the pruning is controlled by the parameters $\lambda$ and $\gamma$, i.e., $\lambda$ regulates the aggressiveness of the pruning, $\gamma$ defines the threshold used to identify the less active neurons of the network (Eq. \ref{eq:gamma}). 
On the other hand, the learning dynamics are controlled by several parameters, but two, i.e., learning rate and batch size, play a very important role. They affect the updates of the model during backpropagation, actively driving the convergence of the learning and determining its final performance. To preserve the possibility of converging to good solutions with a network that progressively loses some of its parts, we adaptively reduce the learning rate to counterbalance the destabilizing effects of the hard pruning process. Alternatively, as it is shown in ~\cite{Smith2018}, it is possible to obtain the same generalization performance by adaptively adjusting the size of the mini-batches during the epoch-by-epoch training.

In this paper, we exploit the equivalence above. We propose to \emph{dynamically update} the size of the mini-batches during the training while keeping fixed the learning rate. By doing so, we obtain a double benefit because with a single parameter to tune (i.e., the growth rate of the mini-batches, as we will explain in the following), we control both the speed of convergence and the total amount of memory used by the learning process. 
We dynamically regulate the batch size according to the \emph{relative variance} (or Variance-To-Mean Ratio) of the gradients. Precisely, we link the size of the mini-batch to the amount of variability (or noise) contained in the gradients. The effect of using this index is to incrementally increase the size of the mini-batch, i.e., to allow for more precise estimations of the gradient, according to the amount of variance, i.e., instability, of the gradients. To compute this index, we use the algorithm proposed in \cite{Balles2017} which gives a simple and effective way to estimate the variance of the gradient after each gradient computation.


Formally, let $S$ be the variance estimation of the gradients for the current mini-batch and $F$ the value of the loss function for the current mini-batch. At the end of each gradient computation, the new size $b$ of the mini-batch is computed as:
\begin{eqnarray}
\label{eq:batch_increment}
b &=& b + \left\lfloor (1 - \alpha_{bs}) \frac{\|S\|_1}{F}\right\rfloor \label{eq:bs}
\end{eqnarray}
where $\alpha_{bs}, \in [0,1]$ is a smoothing parameter used to drive the batch size according to the observed variance on the gradient. Therefore, the parameter $\alpha_{bs}$ provided in Equation \ref{eq:bs} is the only parameter employed in our method to drive the learning process by varying the batch size according to the stability of the gradient estimation. As mentioned before, Equation \ref{eq:batch_increment} shows that our batch size can only increase epoch by epoch by a quantity that is proportional to the variance of the gradient. The rationale is that in this way, we exploit the higher variance of the gradients to accelerate the convergence of SGD to better solutions, contrasting the effect of hard pruning. Moreover, as the training evolves, the increasing mini-batches stabilize the training and fine-tune the network.

A clear effect of the introduction of Equation \ref{eq:bs} is the fact that the batch size can indefinitely grow and, especially in resource-constrained devices, where memory is limited, this effect may harm the training of the neural network. To avoid this effect while keeping this approach exploitable in our fixed memory budget settings, we guarantee that DynHP does not exceed a given memory budget as follows. At the end of each training epoch, after having pruned the network, we compute the memory available for the growth of the $i$-th mini-batch size as the difference between the memory budget ($\mathcal{C}$) and the current memory occupation of the network $\mathcal{N}_i$.

\begin{equation}
\mathcal{B}_i= \mathcal{C} - \mathcal{N}_i 
\end{equation}
We impose the maximum increase of the mini-batch to be
$$
\Delta\mathcal{B}_i=b_i+(\mathcal{C}-\mathcal{N}_i)
$$
Denoting with $\tilde{b}_{i+1}$ the ``candidate'' size of the mini batch, computed through Equation~\ref{eq:bs}, and taking into account the above maximum size limitation, we obtain that the mini-batch size at epoch $i+1$ is:
\begin{equation}
b_{i+1} = \min(\Delta\mathcal{B}_i,\tilde{b}_{i+1}).
\label{eq:batch_cap}
\end{equation}
Note that the network pruning is performed at the end of the epoch; therefore, at the beginning of the next epoch, additional free memory might be available to grow the size of mini-batches, i.e., $\mathcal{B}_{i+1} \geq \mathcal{B}_i$ always holds.


\section{Experiments}
\label{sec:experiments}
The experiments conducted aim to comprehensively evaluate our proposal and compare it with the state-of-the-art SP competitor \cite{Louizos2017}. More precisely, we are interested in answering the following research questions:

\begin{itemize}
	\item[RQ1:] To what extent does our HP technique compresses the network and reduces over-parametrization? How much does HP impact the quality of the learned model?
	\item[RQ2:] Can we reduce the possible loss of accuracy introduced by HP by dynamically adjusting the size of mini-batches during the learning process? To what extent does DynHP impact: i) convergence speed, ii) memory occupation and iii) computational effort?
\end{itemize}

\subsection{Datasets and Competitors}
We validate the performance of our method on the well-known task of multi-class image classification on three public datasets. In detail, we perform i) handwritten digit recognition on the MNIST dataset \cite{lecun1998mnist}, ii) fashion product classification on the Fashion-MNIST dataset \cite{xiao2017/online}, iii) object classification on the CIFAR-10 dataset \cite{krizhevsky2009learning}. Table \ref{tab:datasets} reports some salient statistics of the three datasets employed in this study.

\begin{table}[ht]
	\caption{Properties of the three datasets employed.\label{tab:datasets}}
	\centering
	\begin{tabular}{@{}lcccc@{}}
		\multirow{2}{*}{Dataset} & \multicolumn{2}{c}{\# Images} & \multirow{2}{*}{Image size (\# pixels)} & \multirow{2}{*}{\# channels} \\
		\cline{2-3}
		& Training Set & Test Set & & \\
		\toprule
		MNIST & 60,000 & 10,000 & 28$\times$28 & 1\\
		Fashion-MNIST & 60,000 & 10,000 & 28$\times$28 & 1 \\
		CIFAR-10 & 50,000 & 10,000 & 32$\times$32 & 3 \\
		\bottomrule
	\end{tabular}
\end{table}

For the sake of comparison, we employ the same network architectures, i.e., a Multi-Layer Perceptron (MLP) and a ResNet (RN) \cite{he2016deep} firstly employed by Louizos \emph{et al.} \cite{Louizos2017}. The MLP network is used for the first two tasks that employ MNIST and Fashion-MNIST, while RN is employed on CIFAR-10. The MLP network consists of two hidden layers of size $300$ and $100$ and an output layer of $10$ neurons to deal with multi-class classification.\footnote{Our MLP network achieves comparable results with the publicly available ones on the Fashion-MNIST dataset. See \url{https://github.com/zalandoresearch/fashion-mnist}.} Moreover, the input layer is composed of $784$ neurons as each neuron encodes a pixel of an input image. The RN architecture follows the one originally presented by He \emph{et al.} \cite{he2016deep}, i.e. ResNet-28-1. We adopt the same configuration of the gating mechanism initially set by Louizos \emph{et al.} \cite{Louizos2017}. Precisely, in convolutional layers, the gates are applied to the output channel of the convolutional filters (one gate for each filter), while in dense layers, we have one gate for each neuron. Although there might exist better configurations, for the sake of comparison, we kept the same initially proposed by Louizos \emph{et al.} for SP \cite{Louizos2017}.

We compare the performance of both HP and DynHP presented in Sections~\ref{sec:HP} and  ~\ref{sec:dynhp} against SP. We perform hyper-parameter tuning for both networks and report the results of the best combination of parameters found for each of the three techniques. For the MLP network, in both HP and DynHP, we set the parameter $\lambda$ that controls the aggressiveness of the pruning process to $0.01$, the pruning threshold $\gamma$ to $0.5$ and, we train it for $2000$ epochs. For the RN architecture, we set $\lambda=10^{-5}$, $\gamma=0.47$ and we train it for $200$ epochs. SP follows the configuration of the original paper by Louizos \emph{et al.} \cite{Louizos2017}. Regarding the parameter $\alpha_{bs}$ we perform a sensitivity analysis, and we present the results in Sections~\ref{sub:RQ1_results} and \ref{sub:RQ2_results}.

In the experimental results, we show the capabilities of HP and DynHP in saving memory compared to SP, and to what extent the pruning process affects the final model's accuracy.

\paragraph*{Evaluation metrics}
We measure the generalisation capabilities of the network models learned by using the misclassification error rate $$ERR = \frac{1}{N_{te}}\sum_{i=1}^{N_{te}}I_{\hat{y}_i\ne y_i}$$ where $N_{te}$ is the cardinality of the test set.  Moreover, to understand the impact of each technique on Memory Occupation (MO), we analyze the size of the memory occupation during training. MO is computed as the integral over time (over epochs) of memory usage.
Precisely, MO is computed as the sum of the non-zero (Z) floating-point numbers of the weight matrices of each layer of the neural network plus the ones used to store a mini-batch of the data. In our implementations, we adopt a $32$-bit representation for floating-point.
We also measure the computational effort required to perform inference on the final trained models in terms of the number of floating-point operations (FLOPs).

\subsection{Impact of Hard Pruning (RQ1)}
\label{sub:RQ1_results}
We start the analysis of our hard pruning method by evaluating its impact on the entire learning process. In this first set of results, we use the three datasets to compare Hard Pruning and Soft Pruning on two aspects: i) the effectiveness of the pruning process and ii) the accuracy of the final solution.

\begin{figure}[ht]
    \centering
    \begin{subfigure}[b]{0.32\textwidth}
        \centering
		\includegraphics[width=1\textwidth]{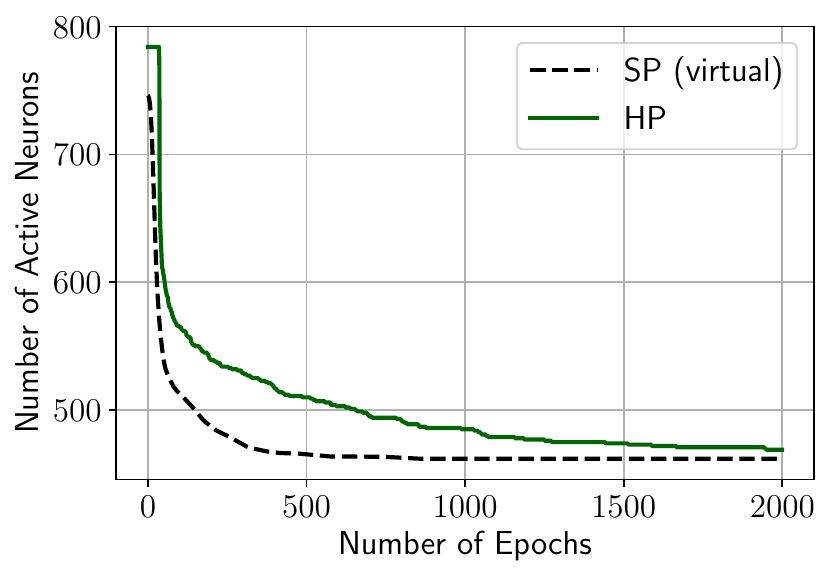}
		\caption{Input Layer (784 neurons)}
    \end{subfigure}%
    ~
    \begin{subfigure}[b]{0.32\textwidth}
        \centering
        \includegraphics[width=1\textwidth]{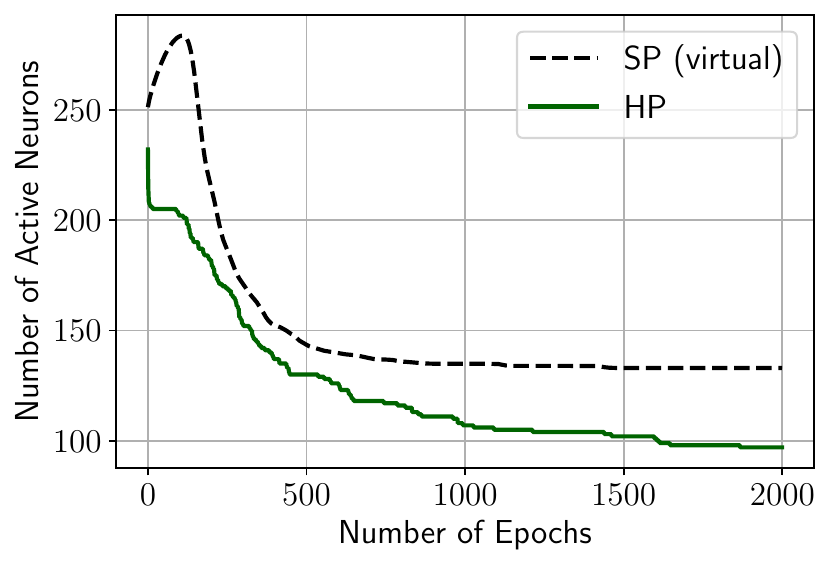}
		\caption{Hidden Layer 1 (300 nodes)}
    \end{subfigure}%
    ~
    \begin{subfigure}[b]{0.32\textwidth}
        \centering
		\includegraphics[width=1\textwidth]{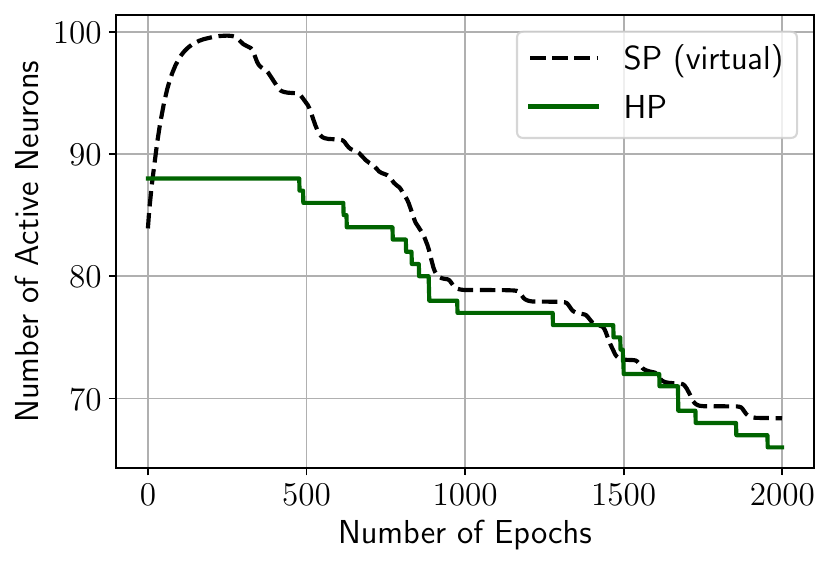}
		\caption{Hidden Layer 2 (100 nodes)}
    \end{subfigure}\\
    \begin{subfigure}[b]{0.32\textwidth}
        \centering
		\includegraphics[width=1\textwidth]{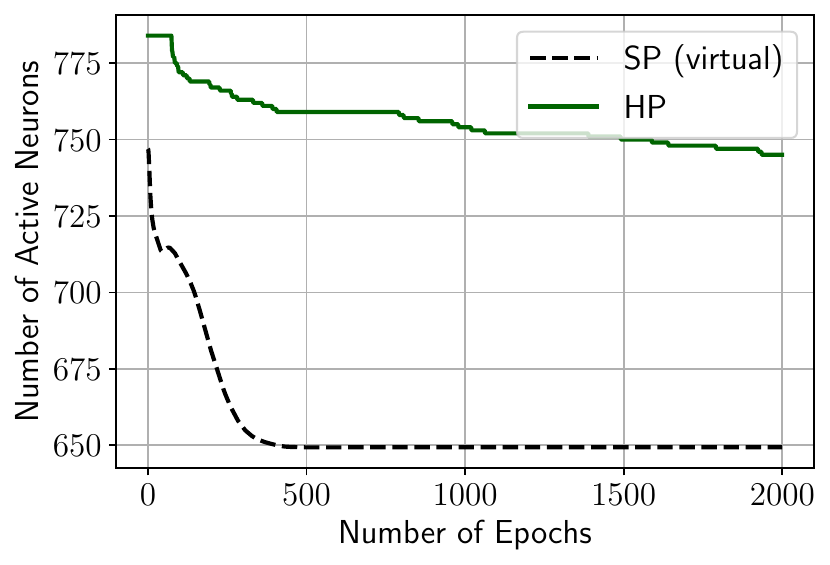}
		\caption{Input Layer (784 nodes)}
    \end{subfigure}%
    ~
    \begin{subfigure}[b]{0.32\textwidth}
        \centering
        \includegraphics[width=1\textwidth]{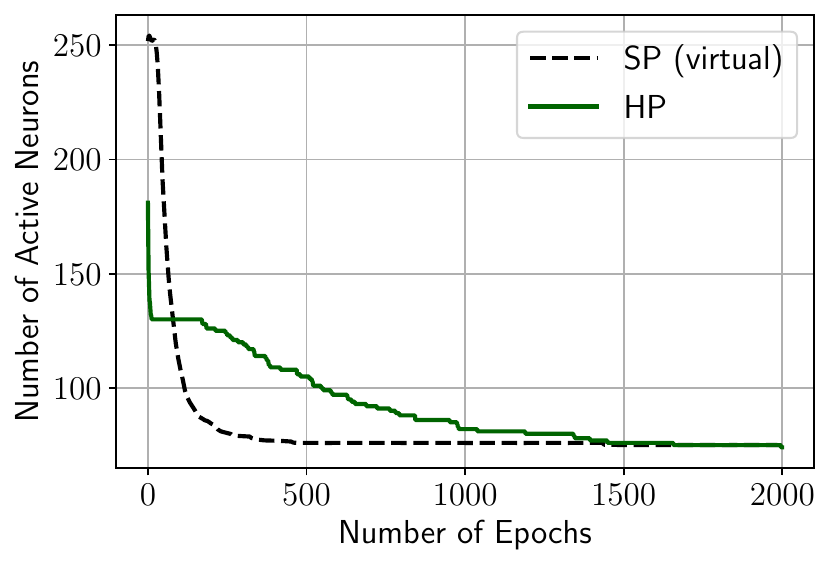}
		\caption{Hidden Layer 1 (300 nodes)}
    \end{subfigure}%
    ~
    \begin{subfigure}[b]{0.32\textwidth}
        \centering
		\includegraphics[width=1\textwidth]{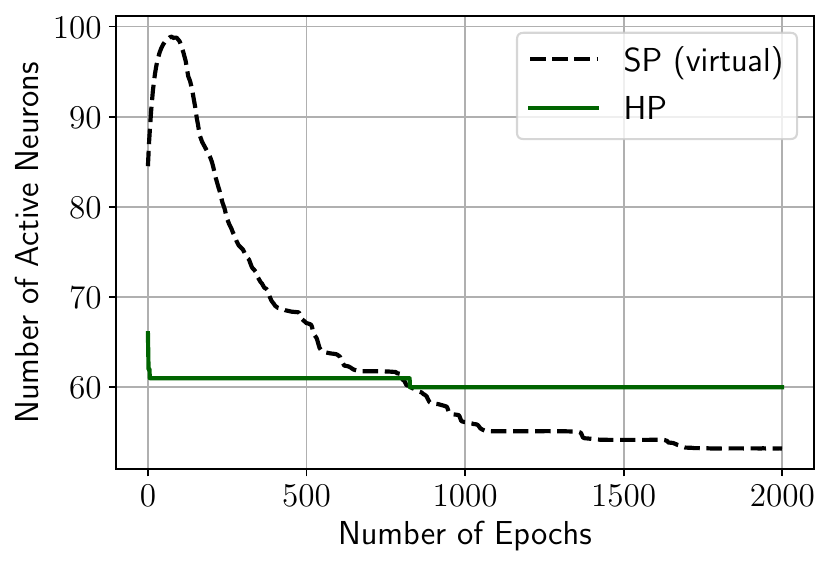}
		\caption{Hidden Layer 2 (100 nodes)}
    \end{subfigure}
	\caption{First row refers to MNIST, second row refers to Fashion-MNIST. The curves show the number of active neurons for SP and HP for each layer of the network, i.e., input and two hidden layers, by increasing the number of epochs of the training.\label{fig:nodes_per_epoch}}
\end{figure}

We evaluate the effectiveness of the pruning process by performing a layer-by-layer analysis of both methods, reporting the different behaviour of SP and HP in reducing the number of active neurons in Figure \ref{fig:nodes_per_epoch}. We report this analysis on the MLP network on the MNIST and Fashion-MNIST datasets. Similar behaviour is achieved for the RN network on CIFAR-10. Precisely, the first row of plots (Figures \ref{fig:nodes_per_epoch}a-c) refers to the MNIST dataset while the plots in Figures \ref{fig:nodes_per_epoch}d-f refer to Fashion-MNIST. Each plot shows the number of active neurons in each layer of the network along with the training. From left to right, we show the first, second and third layer of the network, respectively. For this comparison, we report only the curves of the best operating configurations in terms of accuracy, as will be clear from the results we show later on. For SP the best configurations are those with batch size $512$ for MNIST and $128$ for Fashion-MNIST. For HP we set them to $100$ and $128$, respectively. Moreover, while in HP the number of active neurons corresponds to the actual size of the layers, in the case of SP no neurons are actually pruned, so the size of the network (corresponding to this memory occupation) is (much) higher than the number of active neurons, i.e. in SP the memory occupation during training remains constant. However, to better understand the mechanisms of both SP and HP, we plot for SP the average number of active neurons during the training.

For both datasets, SP starts with a smaller number of active neurons w.r.t. the ones available, e.g., $250$ out of $300$ in hidden layer 1 (HL-1) and $\sim80$ out of $100$ in hidden layer 2 (HL-2) and during the initial training epochs this number increases. In this phase, SP explores several configurations reactivating previously muted neurons. Note that SP at some point reactivates almost all the neurons in HL-2. Then, it starts decreasing again until the end of the training epochs. This behaviour is due to how SP works: if the initial pruning is too aggressive, it adds back neurons to the network. It is worth noting that such behaviour greatly differs between layers and datasets.

\begin{figure}[ht]
	\centering
	\begin{subfigure}[b]{0.4\textwidth}
	\centering
	\includegraphics[width=1\textwidth]{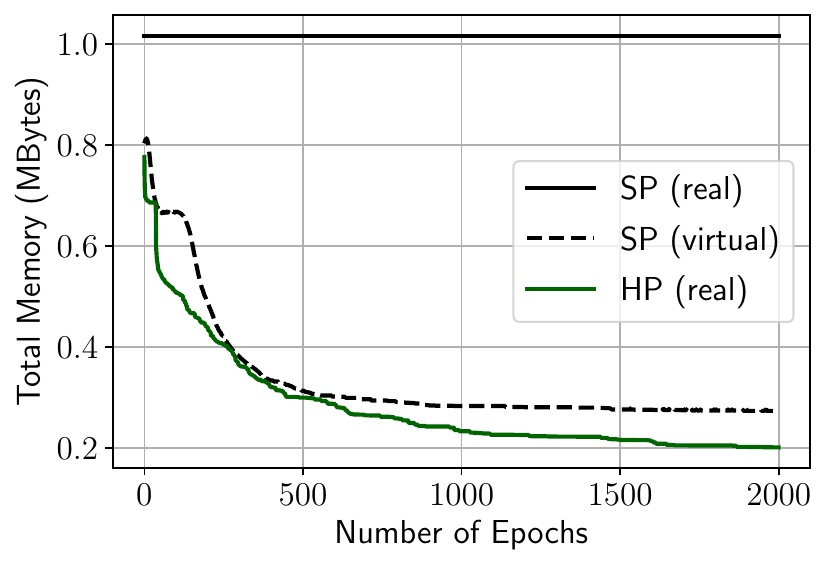}
	\caption{MNIST}
	\end{subfigure}
	~
	\begin{subfigure}[b]{0.4\textwidth}
	\centering
	\includegraphics[width=1\textwidth]{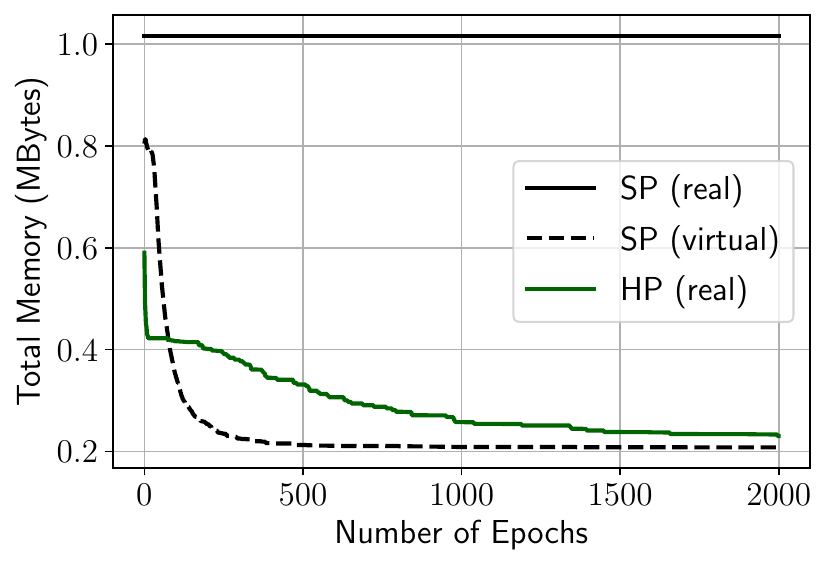}
	\caption{Fashion-MNIST}
	\end{subfigure}
	\caption{Total memory (MBytes) used by SP and HP during the $2000$ epochs of the training phase.\label{fig:total_memory_per_epoch}}
\end{figure}

Conversely, this consideration does not hold for HP, where the number of active neurons monotonically decreases in all the network layers. Depending on the specific dataset, the rate at which the pruning removes the unnecessary neurons changes but, still, we can recognize a clear pattern. Specifically, in HP, once a neuron is removed, it cannot be added back to the network. Moreover, as it will be clear in Section \ref{sub:RQ2_results}, the pruning is, to some extent, connected to the convergence rate of the learning process. The monotonicity of HP's pruning process let us shed light on an important difference between the two methods: while SP always uses all the memory for training the network, HP allows to reduce memory usage by \emph{removing} rows/columns of the weights matrix. We experimentally assess the effect above by analyzing the memory occupation of SP and HP during training. Figure \ref{fig:total_memory_per_epoch} shows the total memory used during the $2000$ training epochs, for both datasets. 

Experiments show that HP, from the very beginning, uses $25\%$ and $45\%$ less memory than SP (solid black line) for MNIST and Fashion-MNIST, respectively. The overall memory savings at the end of the training is $80\%$ and $78\%$ for MNIST and Fashion-MNIST, respectively. It is worth noting that with SP, we can obtain a network of the same size as the one obtained with HP only once the training is completed. Conversely, with HP, we can progressively reduce the size of the network as training progresses. Interestingly, comparing SP (virtual) and HP on MNIST, we notice that the final sizes of the network trained with SP (virtual) and HP are comparable. Coupled with the accuracy results we discuss next, this proves that our method, epoch by epoch, is able to identify the unnecessary neurons correctly, possibly without harming the quality of the training process.

\begin{figure}[ht]
	\centering
	\begin{subfigure}[b]{0.48\textwidth}
	\centering
	\includegraphics[width=1\textwidth]{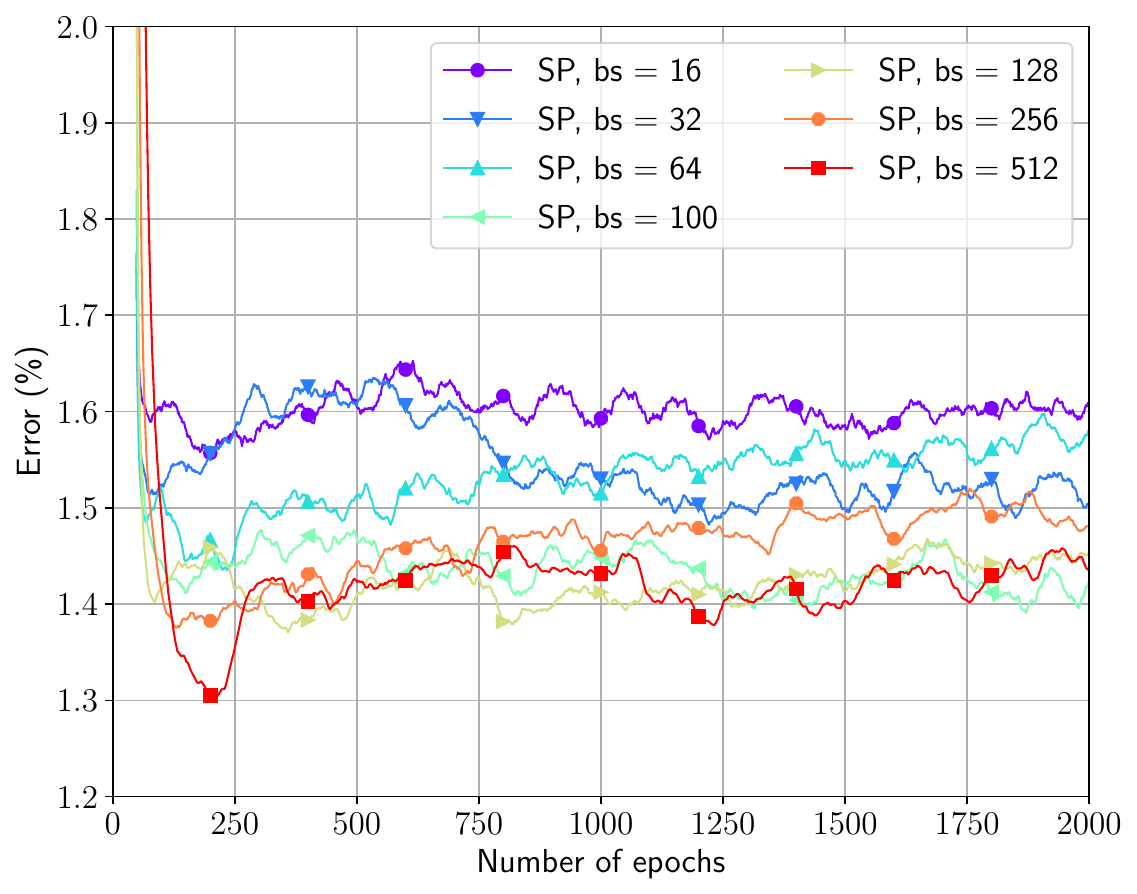}
	\caption{SP\label{fig:mnist_sp_err}}
	\end{subfigure}
	~
	\begin{subfigure}[b]{0.48\textwidth}
	\centering
    \includegraphics[width=1\textwidth]{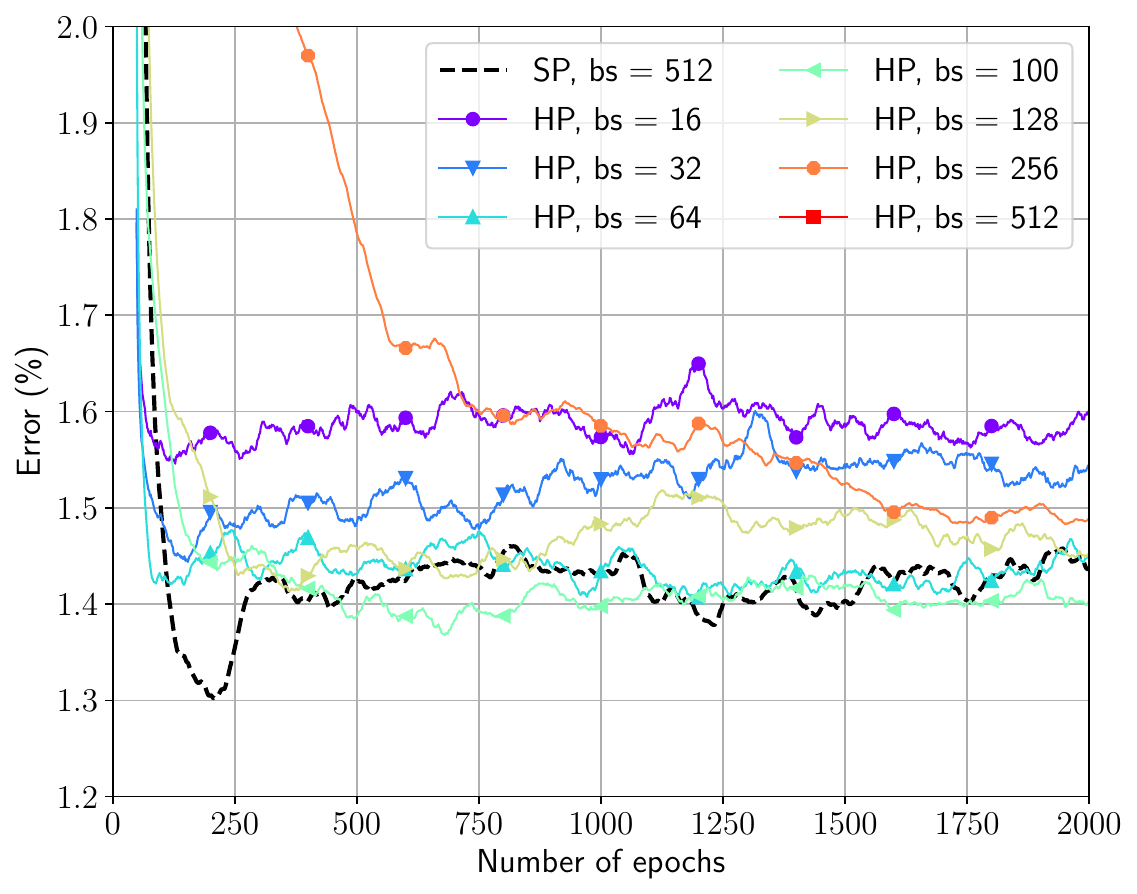}
	\caption{HP vs SP\label{fig:mnist_vhp_err}}
	\end{subfigure}
	\caption{MNIST dataset. Misclassification error ($\%$) of SP and HP for different sizes of mini-batches as a function of the number of epochs of the training.\label{fig:mnist_error_per_epoch}}
\end{figure}

\begin{figure}[ht]
	\centering
	\begin{subfigure}[b]{0.48\textwidth}
	\centering
	\includegraphics[width=1\textwidth]{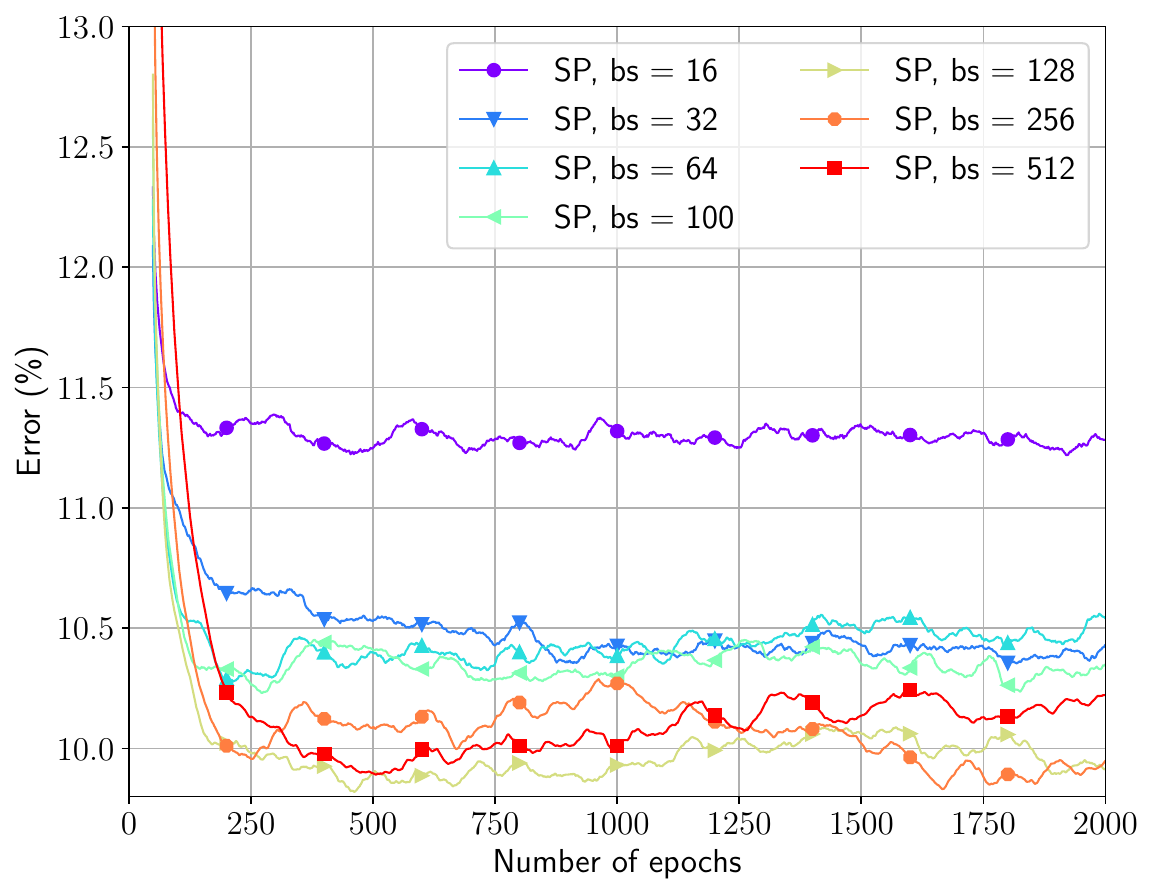}
	\caption{SP\label{fig:fmnist_sp_err}}
	\end{subfigure}
	~
	\begin{subfigure}[b]{0.48\textwidth}
	\centering
	\includegraphics[width=1\textwidth]{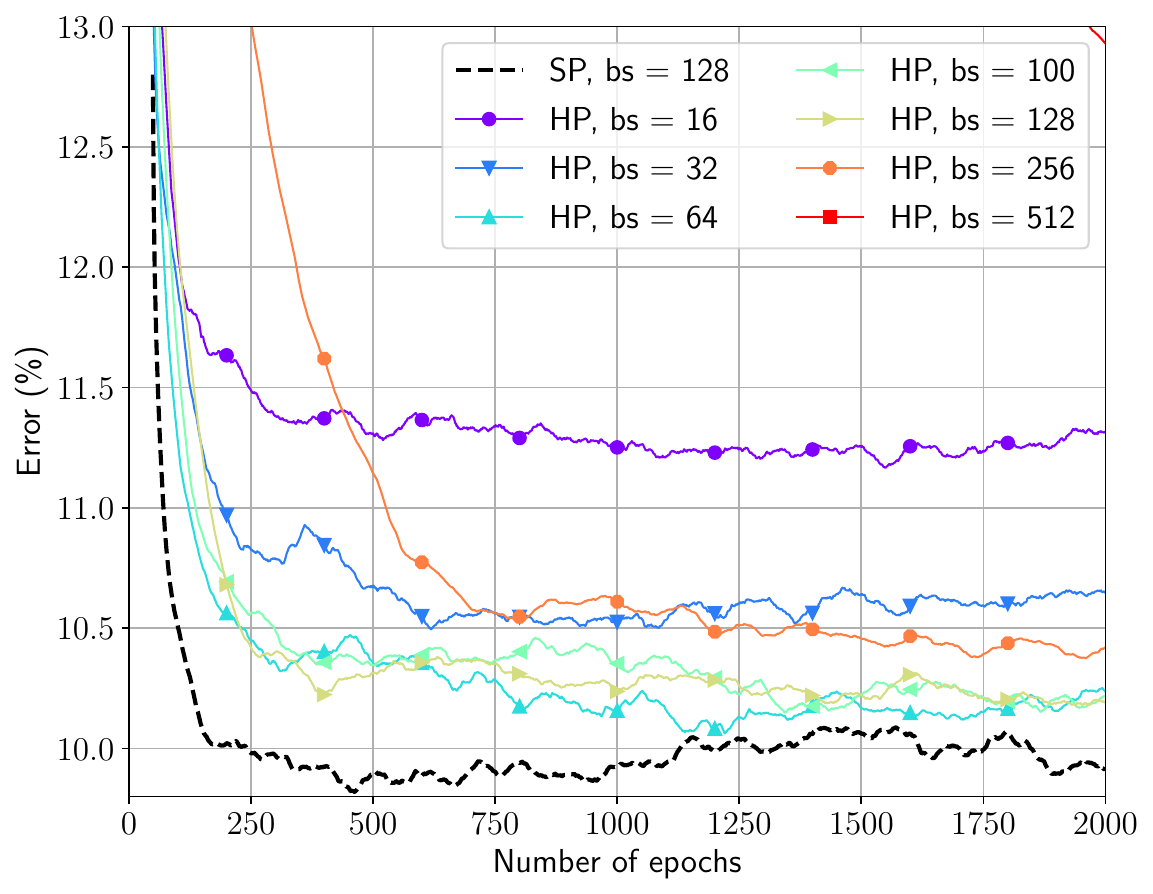}
	\caption{HP vs SP\label{fig:fmnist_vhp_err}}
	\end{subfigure}
	\caption{Fashion-MNIST dataset. Misclassification error ($\%$) of SP and HP for different sizes of mini-batches as a function of the number of epochs of the training.\label{fig:fmnist_error_per_epoch}}
\end{figure}

\smallskip
Let us now evaluate the accuracy of our method. We recall that although the HP mechanism leads to a remarkable saving of memory, it also forbids any possibility to rollback on the pruned neurons that might be needed to regain the learning power of the network ``damaged" by the ablation. Hence, the following analysis compares HP and SP in terms of misclassification error to show the impact of HP on the quality of the final model learned. 
Figures \ref{fig:mnist_error_per_epoch} and \ref{fig:fmnist_error_per_epoch} plot the test error as a function of the number of epochs for sizes of the mini-batches varying in the set $\{16, 32, 64, 100, 128, 256, 512\}$. First, we show in Figures \ref{fig:mnist_sp_err} and \ref{fig:fmnist_sp_err}, for both datasets, the performance of SP at different values of the batch size. As we can see - and as expected - the mini-batch size affects the performance of the network trained with SP. The reason is the interplay between the pruning process and the amount of information controlled by the mini-batch size (the larger the batch size, the more information is usable for a single update). SP obtains the best performance with batch size $512$ for MNIST and $128$ for Fashion-MNIST. The corresponding average misclassification errors are $1.40$ and $9.96$, respectively. As anticipated, we use those configurations as a reference benchmark for the rest of the comparisons. 
Interestingly, even if SP and HP use the same loss function, the two approaches react differently to the batch size. Looking at Figures \ref{fig:mnist_vhp_err} and \ref{fig:fmnist_vhp_err}, we notice that the minimum error achieved by HP are $1.40\%$ and $10.20\%$ with a batch size of $100$ and $128$ for MNIST and Fashion-MNIST, respectively.  The results show that SP performs better than HP in the presence of larger mini-batches, e.g., $512$. This is an expected result: while SP works by only temporarily switching off some of the neurons during learning, i.e., one muted neuron can be reactivated if its contribution is important for the final prediction, HP switches off neurons by removing them permanently from the network.\footnote{The curve for HP with batch size 512 is so high with respect to the others that it does not enter in the plot} Conversely, in HP, the process of hard pruning provides some instability to the learning in a way that SGD performs better with less precise gradient updates (i.e. with more variance) to keep improving the solution. In fact, in terms of convergence speed, we see that HP is generally slower than SP. Nevertheless, HP's solution in both cases is quite accurate and HP results to be only $4\%$ (on average) less precise than SP despite the remarkable lower memory footprint.

\begin{figure}[h!]
	\centering
	\begin{subfigure}[b]{0.48\textwidth}
	\centering
	\includegraphics[width=1\textwidth]{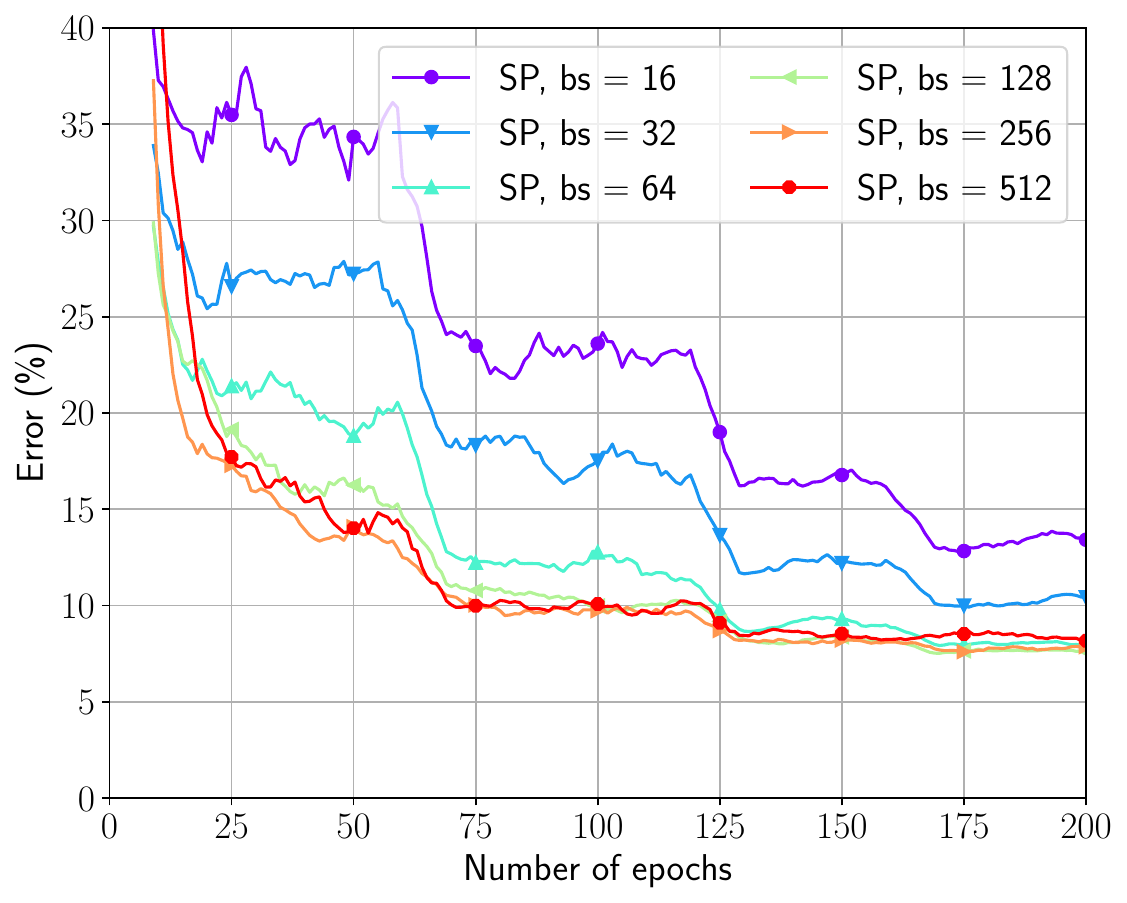}
	\caption{SP\label{fig:cifar10_SP}}
	\end{subfigure}
	~
	\begin{subfigure}[b]{0.48\textwidth}
	\centering
	\includegraphics[width=1\textwidth]{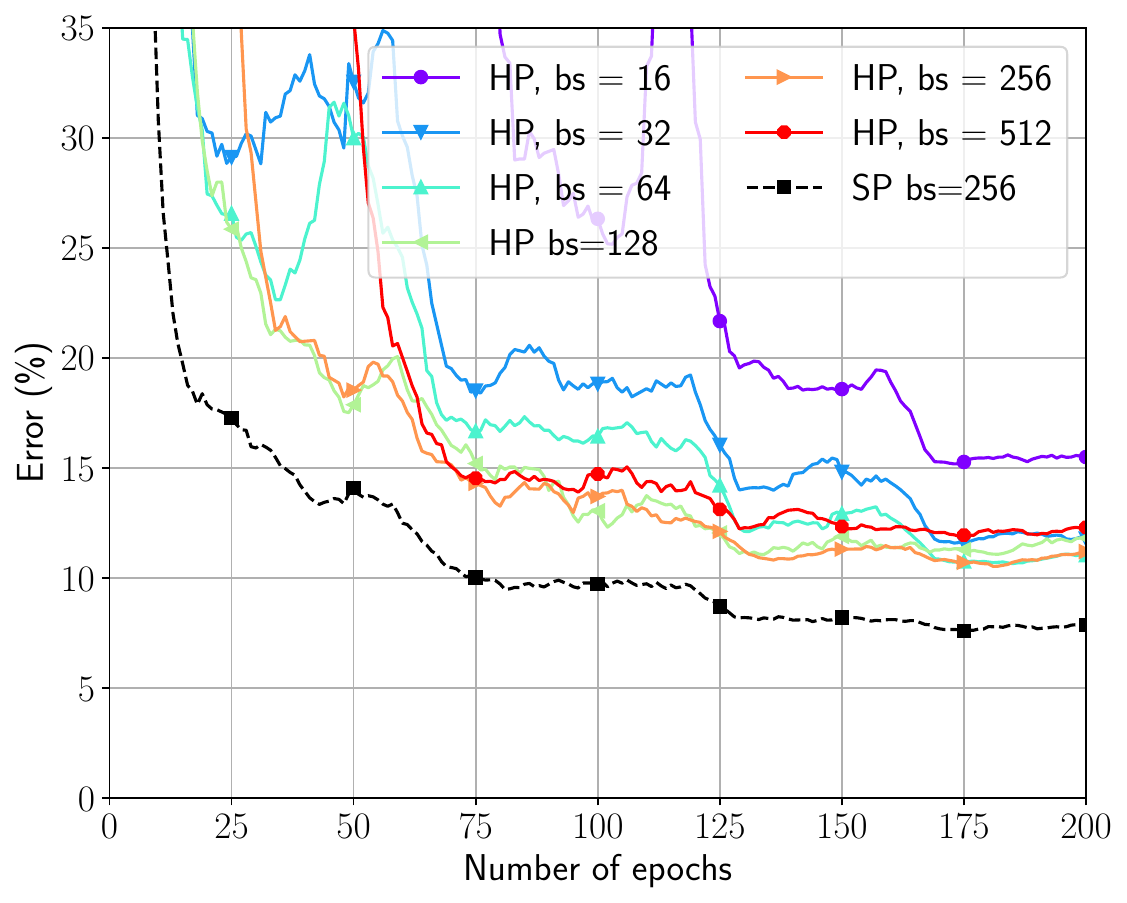}
	\caption{HP vs SP\label{fig:cifar10_SPHP}}
	\end{subfigure}
	
	\caption{CIFAR-10 dataset. Misclassification error ($\%$) of SP and HP for different sizes of mini-batches as a function of the number of epochs of the training.\label{fig:c10_error_per_epoch}}
\end{figure}

The same considerations can be done for the ResNet network on the CIFAR-10 dataset. Figure \ref{fig:cifar10_SP} reports the performance of the network in terms of misclassification error achieved by SP on the test set by varying the batch size during $200$ epochs of training. The best performance corresponds to the configuration with batch size $256$ that allows a final misclassification error of $7.84$\%. On the other hand, Figure \ref{fig:cifar10_SPHP} reports the performance of the network in terms of misclassification error on the test set of HP in the same experimental condition, i.e., by varying the batch size on the same epochs of training. The results of HP are compared against the best performance of SP achieved, i.e., the one with a batch size of $256$. Here, the best performance achieved by HP is the one that exploits a batch size of $256$.
In the case of ResNet, HP achieves a misclassification error of $11.12$\%, increasing the loss achieved by SP by $3.3\%$ only. In the next sections, we further confirm that the loss in accuracy observed for HP is counterbalanced by a significant saving of memory needed to store the network, also in the case of CIFAR-10.

\subsection{Analysis of Dynamic Hard Pruning (RQ2)}
\label{sub:RQ2_results}
From the results shown in Section~\ref{sub:RQ1_results}, it emerges that the size of the mini-batch is a  crucial parameter that must be fine-tuned. However, in a resource-constrained scenario, where computational resources are to be spared, performing the fine-tuning of the batch size might be prohibitive since it would impose to train the network multiple times. In the following, we show the results obtained by introducing our mechanism for dynamically controlling the size of the mini-batches during training. 
In particular, we show that by dynamically adjusting the batch size, we can improve the effectiveness and convergence speed of HP. Moreover, we also evaluate the effect of such a dynamic mechanism on the amount of memory required to train the network.

To make a fair comparison between our DynHP and the other competitors, i.e., SP and HP, in which the size of the mini-batches is fixed a priori (tuned to their best performance), we adopt the following procedure. In Section \ref{sec:dynhp}, we described a mechanism that allows the training process of DynHP not to exceed a fixed memory budget. To this end, we instrument our code to monitor during network training the memory occupation of the network and the one of the mini-batch in the current epoch.\footnote{These quantities dominate the whole memory occupation of the training process.} To fairly compare DynHP with SP, we set the maximum available memory budget of DynHP to the quantity of memory required by the best configuration of SP. In this way, we are sure that DynHP uses, at most, the same amount of memory resources used by SP.

Tables \ref{tab:mnist_memtot} and \ref{tab:fmnist_memtot} report for each method and configuration (i.e., batch size or $\alpha_{bs}$) on the two datasets: i) the final misclassification error with, in parenthesis, the difference (\%) of misclassification error with respect to SP: a negative difference of error indicates an accuracy improvement, while a positive one indicates an accuracy degradation, ii) the final size of the model with, in parenthesis, the percentage of space saved with respect to SP, and iii) the total memory occupation (denoted as ``Tot. Memory Usage'' in the tables) during the training computed as the area under the training-memory curve (like the ones shown in Figure~\ref{fig:dynhp_totmem}) with, in parenthesis, the percentage of memory savings w.r.t. SP based on the whole training epochs. We use this metric to make a fair comparison with SP, i.e. in our method, the total memory usage changes during training, while with SP, it remains fixed. Table~\ref{tab:mnist_memtot} reveals that, with a proper tuning (batch size = 100), on the MNIST dataset, HP achieves $88$\% memory saving on the model and a reduction of memory used during training of up to $77\%$ while slightly improving the accuracy performance ($0.01\%$) w.r.t. SP at the same time. Conversely, in the case of Fashion-MNIST (Table~\ref{tab:fmnist_memtot}), we see that HP achieves a final $77$\% model reduction by using, during training, $52\%$ less memory than SP. In this case, we note an accuracy degradation with respect to SP, which is limited to only $2\%$. These results can be considered an additional confirmation that neural networks are often over-parametrized and, thus, it is possible to significantly reduce their size at a minimal cost, if any, in terms of accuracy.

Let us focus on the results of DynHP where the size of the mini-batches is adapted according to the evolution of the training process.
On the MNIST dataset (Table~\ref{tab:mnist_memtot}), we noticed, in some cases (e.g., $\alpha_{bs}=0.972$) a further increase of accuracy with respect to SP, which is, however, paid with a lower memory saving (with respect to the one obtained with HP) over SP. Moreover, other configurations (e.g., $\alpha_{bs}=0.975$) result in higher memory saving during training, at the cost of a small reduction of accuracy. More in detail, we notice a \emph{trade-off} between the final misclassification error and the final network size. Incrementing $\alpha_{bs}$ the relative error shows a non-linear trend (i.e., it first decreases and then increases) that has its minimum at $\alpha_{bs}=0.972$ where DynHP is 0.06\% and 0.07\% more accurate than  HP and SP, respectively. 
Regarding the relation between $\alpha_{bs}$ and the memory savings, there is a positive correlation, i.e., the higher is $\alpha_{bs}$, the more memory we can save. This holds for both the compression of the model (``Model Size'' column), which ranges from 45\% ($\alpha_{bs}=0.971$) to 83\% ($\alpha_{bs}=0.975$), and for the training memory occupation (``Tot. Memory Usage'' column) where DynHP saves up to 81\% w.r.t. SP. Interestingly, the highest compression (85\% at $\alpha_{bs}=0.975$) does not lead to the best accuracy (1.35 at $\alpha_{bs}=0.972$), which is obtained with a sensibly less compressed model (62\%). 

\begin{table}[h]
	\centering
	\caption{Performance and impact on memory occupation for SP, HP and DynHP on the MNIST dataset. In parenthesis, we report the difference (\%) in terms of misclassification error achieved by HP/DynHP with respect to SP. We also report the space saved (\%) in terms of final model size and memory used for the training achieved by HP/DynHP with respect to SP.\label{tab:mnist_memtot}}
	\begin{tabular}{@{}lllll@{}}
		\multirow{2}{*}{Method} & \multirow{2}{*}{$\alpha_{bs}$} & Misclassification Error & Model Size & Tot. Memory Usage \\
		& & (\%) & (MBytes) & (GBytes) \\
		\toprule
		SP ($bs = 512$)   & -- & $1.42$ (--) & $1.041$ (--) & $5.219$ (--) \\
		HP ($bs = 100$)   & -- & $1.41$ ($-0.01$\%) & $0.206$ ($-88$\%) & $1.185$ ($-77$\%) \\
		\midrule
		DynHP             & $0.971$ & $1.50$ ($+0.08$\%) & $0.572$ ($-45$\%) & $5.165$ ($-1$\%) \\
		DynHP             & $0.972$ & $1.35$ ($-0.07$\%) & $0.398$ ($-62$\%) & $2.736$ ($-48$\%) \\
		DynHP             & $0.973$ & $1.40$ ($-0.02$\%) & $0.205$ ($-80$\%) & $1.279$ ($-75$\%) \\
		DynHP             & $0.974$ & $1.45$ ($+0.03$\%) & $0.217$ ($-79$\%) & $1.361$ ($-74$\%) \\
		DynHP             & $0.975$ & $1.43$ ($+0.01$\%) & $0.178$ ($-83$\%) & $0.980$ ($-81$\%) \\
		\bottomrule
	\end{tabular}
\end{table}

On the Fashion-MNIST dataset (Table~\ref{tab:fmnist_memtot}), at the general level, we notice a similar behaviour as observed with MNIST. Specifically, depending on $\alpha_{bs}$ we can obtain an accuracy improvement with respect to HP at the cost of higher memory usage (particularly during training) or a significant memory saving at the cost of a slight loss of accuracy. At a more granular level, however, we notice a different trend with respect to what is observed with MNIST. First, DynHP with $\alpha_{bs}\in[0.979,0.982]$ performs equally or even better than SP showing up to 0.19\% of accuracy improvement with a model that is 60\% smaller than SP ($\alpha_{bs}=0.980$). However, the memory occupation during training (``Tot. Memory Usage'' column) is equivalent to SP. This behaviour relies on the fact that DynHP, to counterbalance the loss of the network expressiveness, increases the size of the mini-batches. Second, results for $\alpha_{bs}\in[0.987,0.989]$ reveal that under the condition of accepting a small degradation of accuracy (from $0.12\%$ to $0.54\%$ w.r.t. SP), DynHP produces highly compressed models, i.e., up to $88$\% space-saving with up to $79\%$ of memory occupation during training.

\begin{table}[h]
\centering
\caption{Accuracy performance and impact on memory occupation for SP, HP, and DynHPon the Fashion-MNIST dataset. In parenthesis, we report the difference (\%) in terms of misclassification error achieved by HP/DynHP with respect to SP. We also report the space saved (\%) in terms of final model size and memory used for the training achieved by HP/DynHP with respect to SP.\label{tab:fmnist_memtot}}
\small
\begin{tabular}{@{}lllll@{}}
		\multirow{2}{*}{Method} & \multirow{2}{*}{$\alpha_{bs}$} & Misclassification Error & Model Size & Tot. Memory Usage \\
		& & (\%) & (\%) & (GBytes) \\
		\toprule
		SP ($bs = 128$) & -- & $9.96$ (--) & $1.041$ (--) & $2.866$ (--)  \\
		HP ($bs = 128$) & -- & $10.20$ ($+0.24$\%) & $0.236$ ($-77$\%) & $1.377$ ($-52$\%)  \\
        \midrule
		DynHP & $0.979$ & $9.97$  ($+0.01$\%)  & $0.430$ ($-59$\%) & $2.874$ ($0$\%) \\
		DynHP & $0.980$ & $9.64$  ($-0.32$\%)  & $0.415$ ($-60$\%) & $2.874$ ($0$\%) \\
		DynHP & $0.981$ & $9.96$  ($-0.00$\%)     & $0.418$ ($-60$\%) & $2.874$ ($0$\%) \\
		DynHP & $0.982$ & $9.77$  ($-0.19$\%)  & $0.399$ ($-62$\%) & $2.874$ ($0$\%) \\
		DynHP & $0.983$ & $10.21$ ($+0.25$\%)  & $0.407$ ($-61$\%) & $2.872$ ($0$\%) \\
		DynHP & $0.984$ & $10.12$ ($+0,16$\%)  & $0.405$ ($-61$\%) & $2.872$ ($0$\%) \\
		DynHP & $0.985$ & $10.06$ ($+0.10$\%)  & $0.410$ ($-61$\%) & $2.872$ ($0$\%) \\
		DynHP & $0.986$ & $9.93$  ($-0.03$\%)  & $0.386$ ($-63$\%) & $2.858$ ($0$\%) \\
		DynHP & $0.987$ & $10.08$ ($+0.12$\%)  & $0.252$ ($-76$\%) & $2.573$ ($-10$\%) \\
		DynHP & $0.988$ & $10.23$ ($+0.27$\%)  & $0.135$ ($-87$\%) & $0.829$ ($-71$\%) \\
		DynHP & $0.989$ & $10.50$ ($+0.54$\%)  & $0.124$ ($-88$\%) & $0.611$ ($-79$\%) \\
        \bottomrule
\end{tabular}
\end{table}

For what regards ResNet on the CIFAR-10 dataset (Table~\ref{tab:c10_memtot}), the general behaviour only shows one of the features observed in MNIST and Fashion-MNIST. Specifically, in this case, DynHP always slightly sacrifices accuracy with respect to HP to significantly improve in terms of memory savings. At a granular level, we observe interesting trade-offs between misclassification error and total memory usage. First, the best performance observed by DynHP is obtained with $\alpha_{bs} = 0.79$, where the misclassification error achieved is $11.13$\%. This error is higher than the error achieved by SP's best configuration, yielding an additional loss of $3.29\%$. The performance of DynHP is in line with the one of the best HP that achieves a misclassification error of $11.12$\% with a $bs = 256$. However, while HP achieves its best misclassification error with a reduction of the model size of $9$\% with respect to SP, DynHP with $\alpha_{bs} = 0.79$ achieves a much higher reduction of the model size of $30$\%, with a reduction of the total memory usage of $18\%$. On the other hand, when using $\alpha_{bs} = 0.89$, the misclassification error achieved by DynHP is $11.48$\%, i.e., an increase of loss of $3.64\%$ with respect to SP. However, in this configuration, DynHP reduces the model size of up to $33$\% and a total memory usage reduction of $41$\% w.r.t SP.

\begin{table}[h]
	\centering
	\caption{Accuracy performance and impact on memory occupation for SP, HP, and DynHP on the CIFAR-10 dataset. In parenthesis, we report the difference (\%) in terms of misclassification error achieved by HP/DynHP with respect to SP. We also report the space saved (\%) in terms of final model size and memory used for the training achieved by HP/DynHP with respect to SP.\label{tab:c10_memtot}}
	\small
	\begin{tabular}{@{}lllll@{}}
			\multirow{2}{*}{Method} & \multirow{2}{*}{$\alpha_{bs}$} & Misclassification Error & Model Size & Tot. Memory Usage \\
			& & (\%) & (\%) & (MBytes) \\
			\toprule
			SP ($bs = 256$) & -- & $7.84$ (--) & $1.41$ (--) & $882.00$ (--)  \\
			HP ($bs = 256$) & -- & $11.12$ ($+3.58$\%) & $1.28$ ($-9$\%) & $856.70$ ($-3$\%)  \\
			\midrule
			DynHP & $0.71$ & $15.46$  ($+7.62$\%)  & $1.02$ ($-28$\%) & $756.53$ ($-14$\%) \\
			DynHP & $0.73$ & $14.36$  ($+6.52$\%)  & $1.00$ ($-29$\%) & $780.16$ ($-12$\%) \\
			DynHP & $0.75$ & $11.21$  ($+3.37$\%)  & $1.01$ ($-28$\%) & $799.75$ ($-9$\%) \\
		    DynHP & $0.77$ & $13.20$  ($+5.35$\%)  & $0.99$ ($-30$\%) & $763.14$ ($-13$\%) \\
			DynHP & $0.79$ & $11.13$  ($+3.29$\%)  & $0.98$ ($-30$\%) & $726.93$ ($-18$\%) \\
			DynHP & $0.81$ & $12.02$  ($+4.18$\%)  & $0.99$ ($-30$\%) & $749.39$ ($-15$\%) \\
			DynHP & $0.83$ & $11.42$  ($+3.58$\%)  & $0.97$ ($-31$\%) & $693.06$ ($-21$\%) \\
			DynHP & $0.85$ & $12.36$  ($+4.52$\%)  & $0.97$ ($-31$\%) & $637.03$ ($-28$\%) \\
			DynHP & $0.87$ & $12.16$  ($+4.32$\%)  & $0.95$ ($-33$\%) & $544.30$ ($-38$\%) \\
			DynHP & $0.89$ & $11.48$  ($+3.64$\%)  & $0.94$ ($-33$\%) & $522.12$ ($-41$\%) \\
			\bottomrule
	\end{tabular}
\end{table}

Summarising this set of results, we can conclude that in general, both HP and DynHP are able to drastically cut memory usage, either during training or at the end of training (on in both cases), with respect to SP. This is paid, sometimes, not always, with a loss of reduction in the order of a few percentage points. DynHP shows some advantages over HP due to the possibility of dynamically adjusting the mini-batch size. In general, it can further reduce memory usage, at the cost of a small loss of accuracy, in the order of a few per cent. In some cases, namely, MNIST and Fashion-MNIST, it further \emph{improves} accuracy, reducing \emph{at the same time} memory usage. Finally, note that, even when the memory used during training by DynHP is of the same size as SP, the final size of the model is always quite smaller.

\subsubsection{Convergence speed and accuracy}
\begin{figure}[ht!]
	\centering
	\begin{subfigure}[b]{0.48\textwidth}
	\includegraphics[width=1\textwidth]{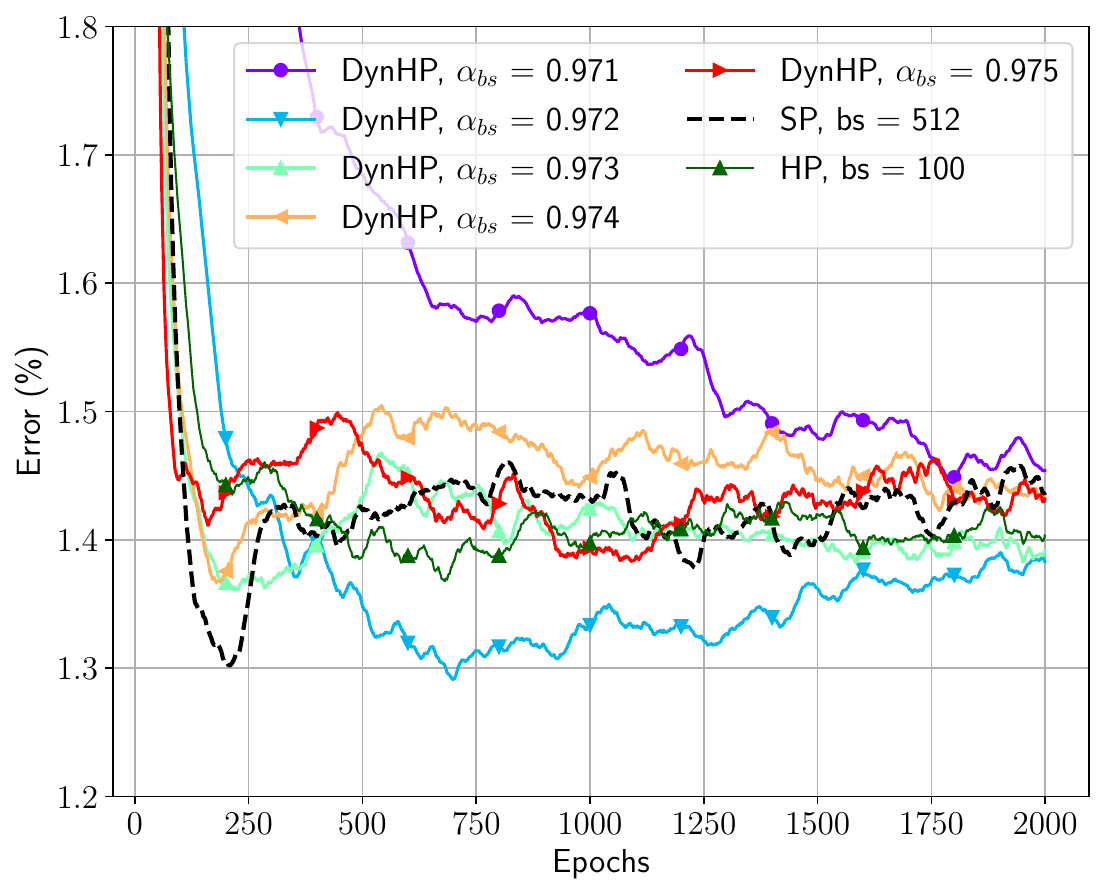}
	\caption{MNIST\label{fig:mnist_dynhp_acc}}
	\end{subfigure}
	~
	\begin{subfigure}[b]{0.48\textwidth}
	\includegraphics[width=1\textwidth]{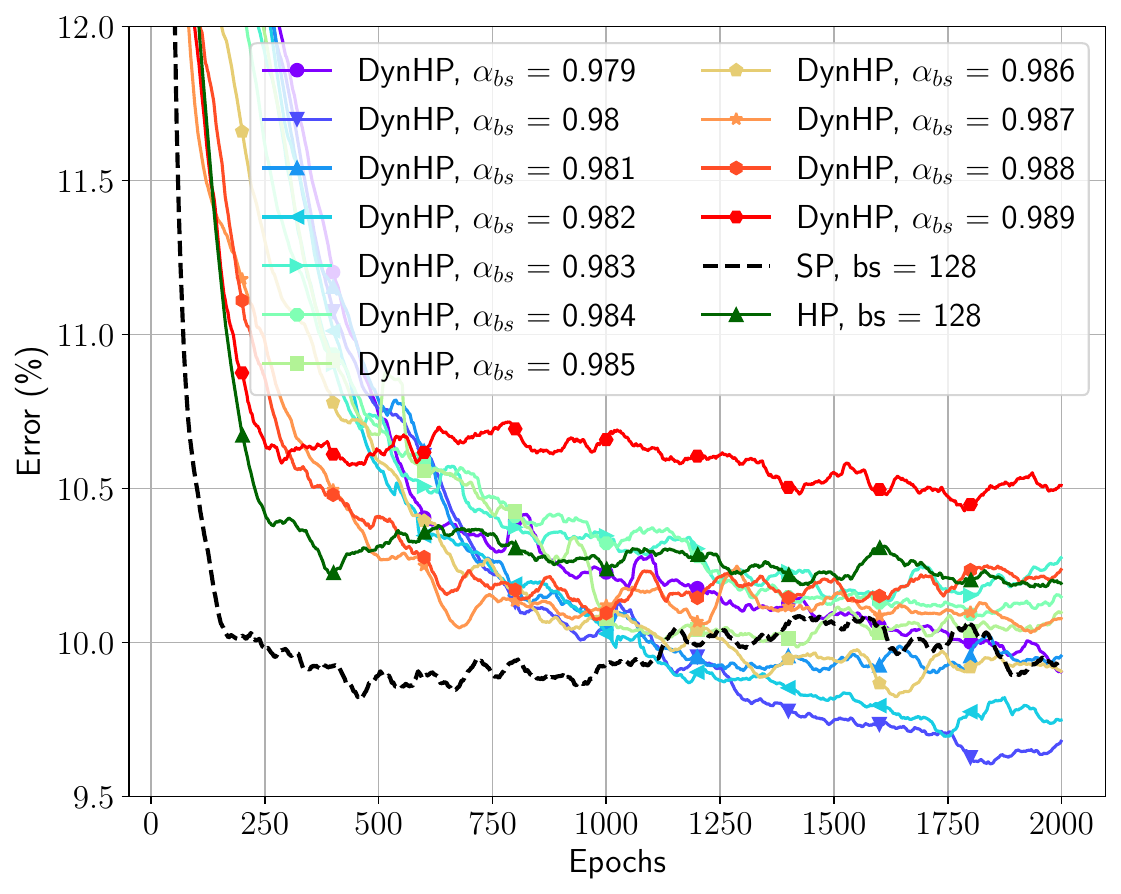}
	\caption{Fashion-MNIST\label{fig:fmnist_dynhp_acc}}
	\end{subfigure}
	~
	\begin{subfigure}[b]{0.48\textwidth}
	\includegraphics[width=1\textwidth]{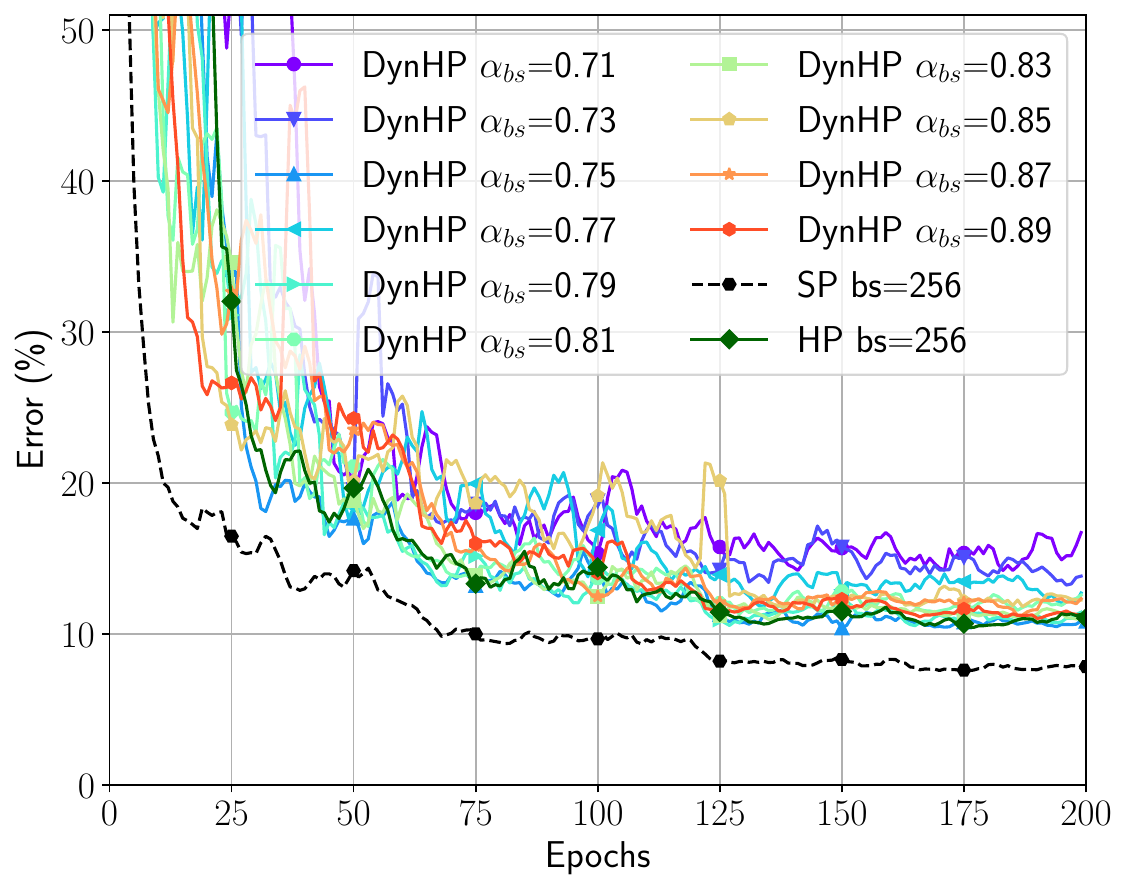}
	\caption{CIFAR-10\label{fig:cifar10-dynhp_acc}}
	\end{subfigure}
	\caption{Misclassification error ($\%$) of SP, HP, and DynHP as a function of the number of training epochs by varying $\alpha_{bs}$.
	\label{fig:dynhp_error}}
\end{figure}

We now go deeper into the analysis by evaluating the convergence speed and the accuracy of SP, HP and DynHP. Figure \ref{fig:dynhp_error} shows the performance on the test set for SP, HP and DynHP. For the sake of clarity and comparison, for SP and HP, we report in the plot only the settings resulting in the best performance, achieved by employing a batch size of 512 and 100, respectively. For DynHP, we report several lines for different values of the parameter $\alpha_{bs}$, responsible for the dynamic increment of the batch size. According to the strategy explained in Section~\ref{sec:dynhp}, we start training with a very small batch size of $16$. 

Looking at Figure~\ref{fig:dynhp_error} we notice that $\alpha_{bs}$ controls the trade-off between training convergence speed and model accuracy. In fact, looking at Figure~\ref{fig:mnist_dynhp_acc}, low values of $\alpha_{bs}$ (e.g., $\alpha_{bs} = 0.971$) induce a quite slow convergence to medium quality solutions while with higher values of $\alpha_{bs}$ (e.g., $\alpha_{bs} = 0.975$)  the results are comparable with SP and HP. Interestingly, for values in between (e.g., $\alpha_{bs} = 0.972$), DynHP converges to a higher quality solution,  outperforming both SP and HP in terms of average misclassification error. As far as Fashion-MNIST is concerned, we notice that the converge speed of DynHP is in general slower that both SP and HP. However, though slower, DynHP shows that for some configurations it is able to out perform both HP and SP.

Figure~\ref{fig:mnist_dynhp_acc} also reports the analysis for the ResNet network on the CIFAR-10 dataset. Here, the optimal performance of DynHP is achieved with $\alpha_{bs} = 0.79$. In this specific configuration, DynHP is able to outperform HP while it achieves a higher loss of 11.13\% with respect to SP. As far as convergence is concerned, in the case of ResNet the convergence of DynHP is not slower than the one of HP and SP even if its trend appears less regular for some specific values of $\alpha_{bs}$.

These results represent a first insight into the potential benefits obtainable by combining dynamic mini-batch size and the hard-pruning. The take-home message is that by fine-tuning the growth rate for the size of mini-batches, we can counterbalance the loss of expressive power of the network due to the irreversible hard-pruning process and converge to better solutions than the ones obtained with SP and HP. Clearly, as the complexity of the network to compress increases, as in RN case, finding the best trade-off between pruning and training becomes more challenging, as reported by the accuracy results.

Finally, note that these findings of convergence are promising, but further investigation is required. In particular, a theoretical derivation of the convergence performance as a function of the algorithm's parameters (and $\alpha_{bs}$ in particular) would be very useful. However, due to the complexity of the required derivations, this is left as future work.

\begin{figure}[ht]
	\centering
	\begin{subfigure}[b]{0.48\textwidth}
	\centering
	\includegraphics[width=1\textwidth]{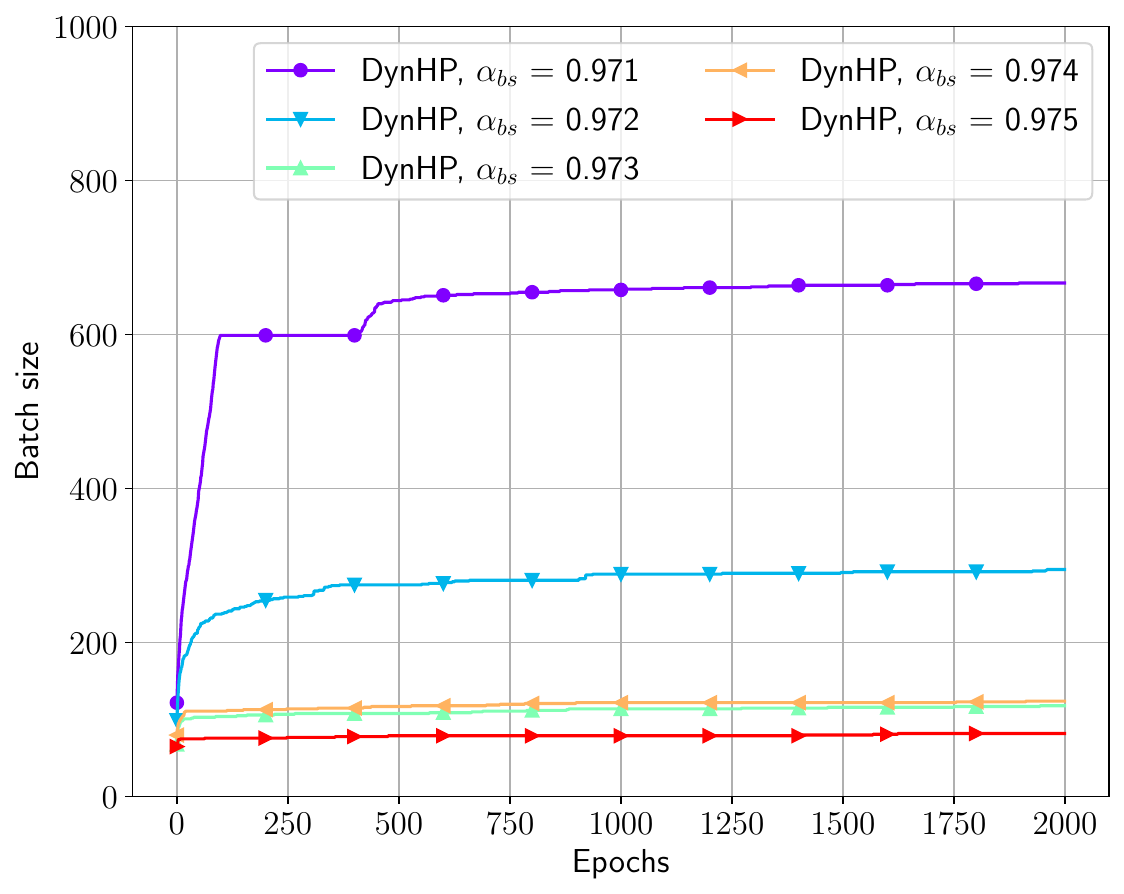}
	\caption{MNIST\label{fig:mnist_dbs}}
	\end{subfigure}
	~
	\begin{subfigure}[b]{0.48\textwidth}
	\centering
	\includegraphics[width=1\textwidth]{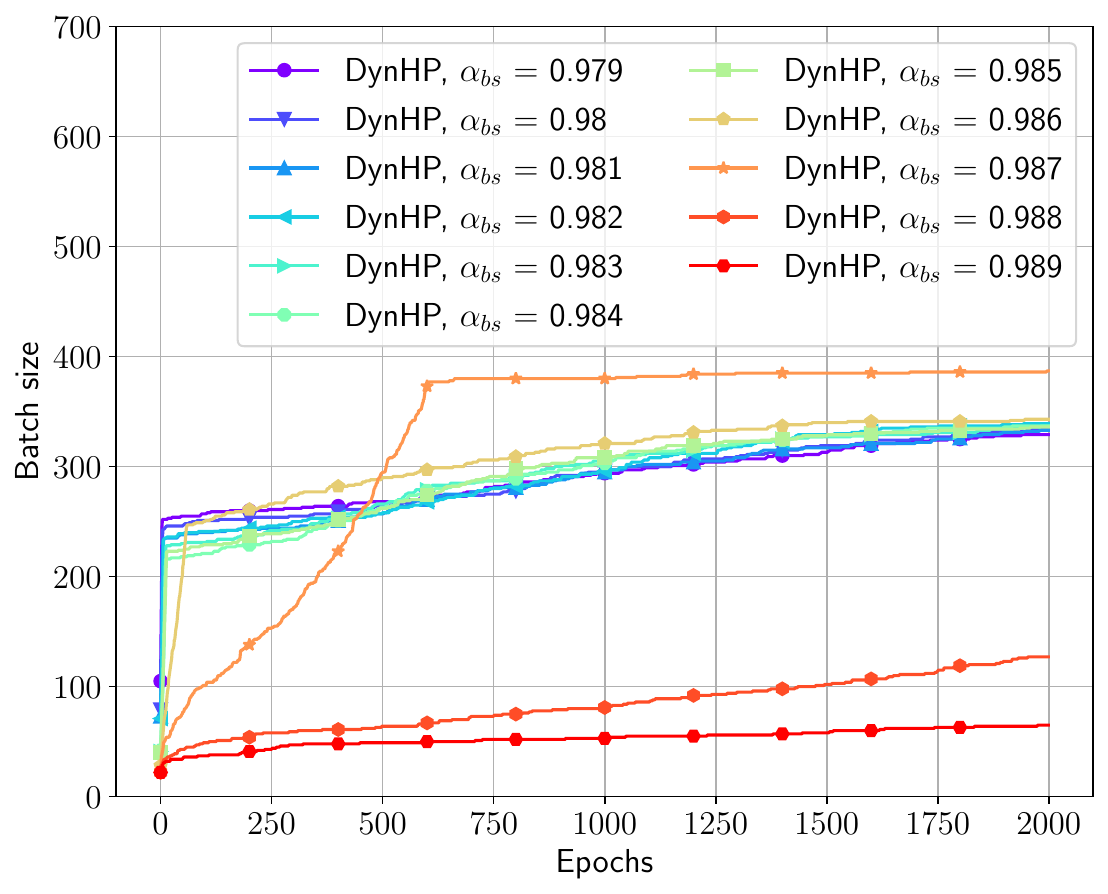}
	\caption{Fashion-MNIST\label{fig:fmnist_dbs}}
	\end{subfigure}
	\\
	\begin{subfigure}[b]{0.48\textwidth}
	\centering
	\includegraphics[width=1\textwidth]{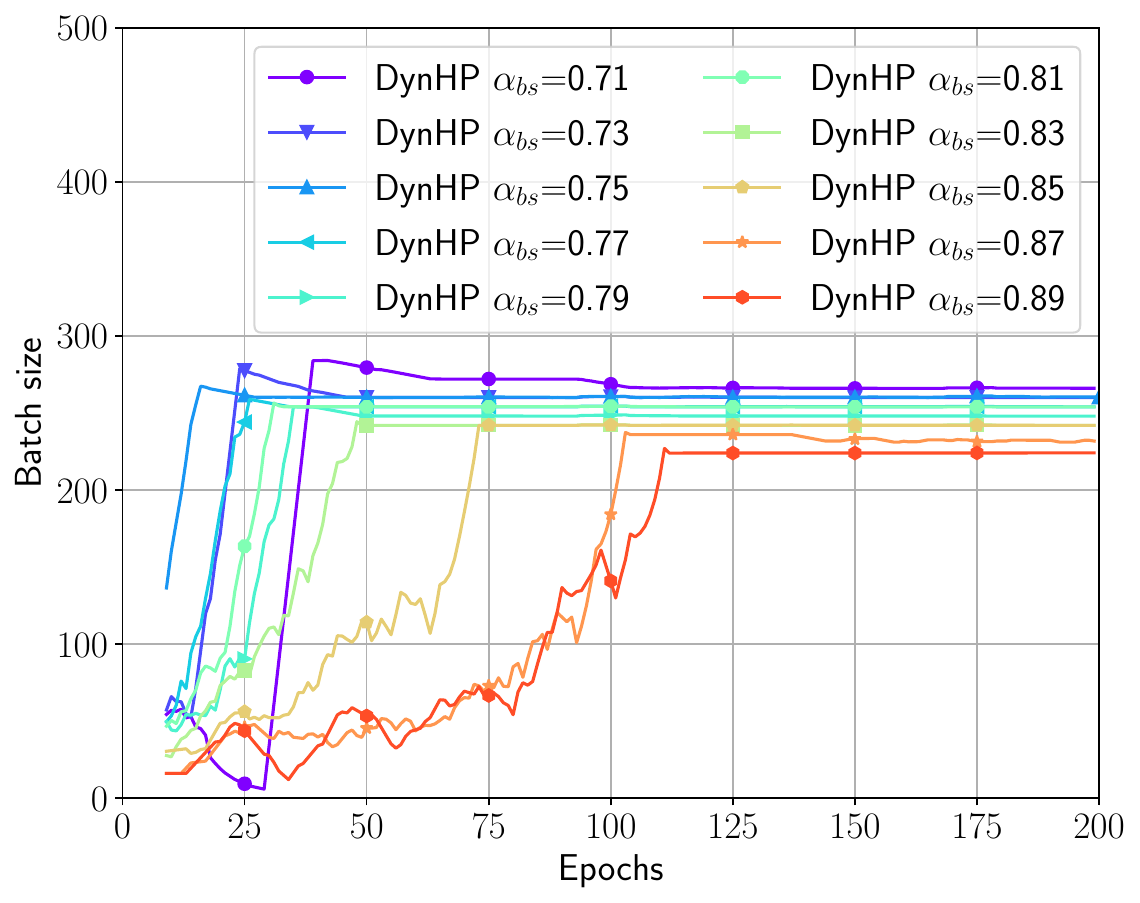}
	\caption{CIFAR-10\label{fig:c10_dbs}}
	\end{subfigure}
	\caption{Size of mini-batches as a function of training epochs for different values of the smoothing parameter $\alpha_{bs}$\label{fig:dynhp_batchsize}.}
\end{figure}

\subsubsection{Memory occupation}
Let us now look at the dynamic batch sizing behaviour and its effect on the memory occupation of the entire learning process. In our configuration, we always start from a small initial batch size (i.e., $bs=16$) and increase it according to relative variance in the gradients. In Figure~\ref{fig:dynhp_batchsize}, we show the size of mini-batches as a function of the training epochs for different values of $\alpha_{bs}$. The first interesting observation regards the fact that the growth of the mini-batches is a convergent process. In fact, for both datasets, we see that almost all the curves stop increasing at different levels during training. Clearly, for each configuration of $\alpha_{bs}$, there is a different size at which the mini-batch stabilizes. It is worth noting that there are two main reasons for which such a memory occupation curve stabilizes. On the one hand, there is the effect induced by Equation~\ref{eq:batch_cap} that forces the size of the mini-batches not to exceed the maximum memory budget. This happens for example for $\alpha_{bs}=0.971$ in Figure~\ref{fig:mnist_dbs} and $\alpha_{bs}=0.987$ in Figure~\ref{fig:fmnist_dbs}. On the other hand, the curve stabilizes when the relative variance of the gradients computed in Equation~(\ref{eq:batch_increment}) becomes negligible, e.g., as for  $\alpha_{bs}=0.972$ in Figure~\ref{fig:mnist_dbs} and $\alpha_{bs}=0.988$ in Figure~\ref{fig:fmnist_dbs}. Qualitatively, the same behaviour can also be observed for CIFAR-10 (Figure~\ref{fig:c10_dbs}).

The second interesting observation is that due to the definition of $\alpha_{bs}$ in Equation~\ref{eq:bs}, the lower the value of $\alpha_{bs}$ the steeper the growth of the batch size.
The effect of such steeper growth is compatible with the convergence performance shown in Figure~\ref{fig:dynhp_error} because, the larger the mini-batches, the slower the convergence.\footnote{Recall that an update is based on the average of the gradients computed on a mini-batch. Therefore, averaging too many different gradients, as they are during the first training epochs, might drive SGD far away from good solutions, increasing the number of iterations needed to converge to better ones.} However, there is no a priori notion of which is the correct steepness that drives the learning process to the best possible result. Our results suggest that the increment of the updates should be small. 

\begin{figure}[ht]
	\centering
	\begin{subfigure}[b]{0.48\textwidth}
	 \centering
	\includegraphics[width=1\textwidth]{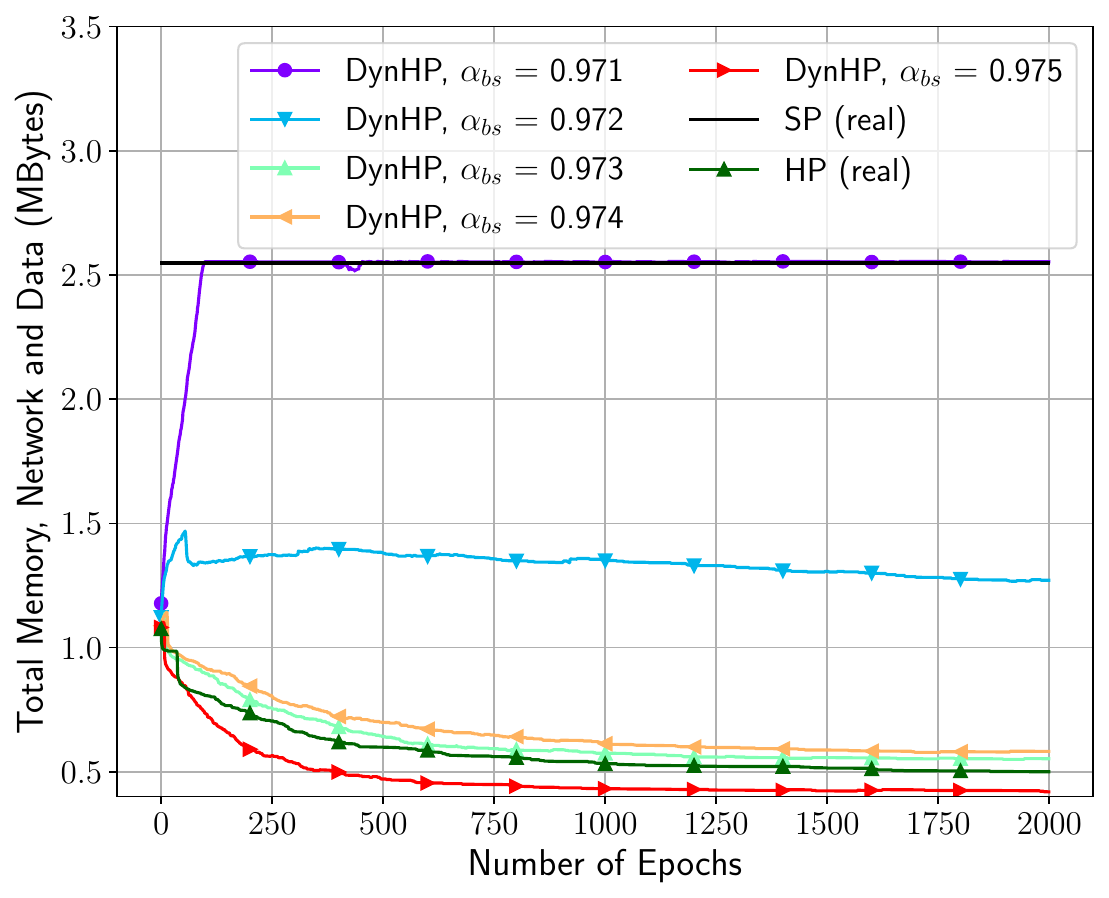}
	\caption{MNIST\label{fig:mnist_totmem}}
	\end{subfigure}
	~
	\begin{subfigure}[b]{0.48\textwidth}
	 \centering
	\includegraphics[width=1\textwidth]{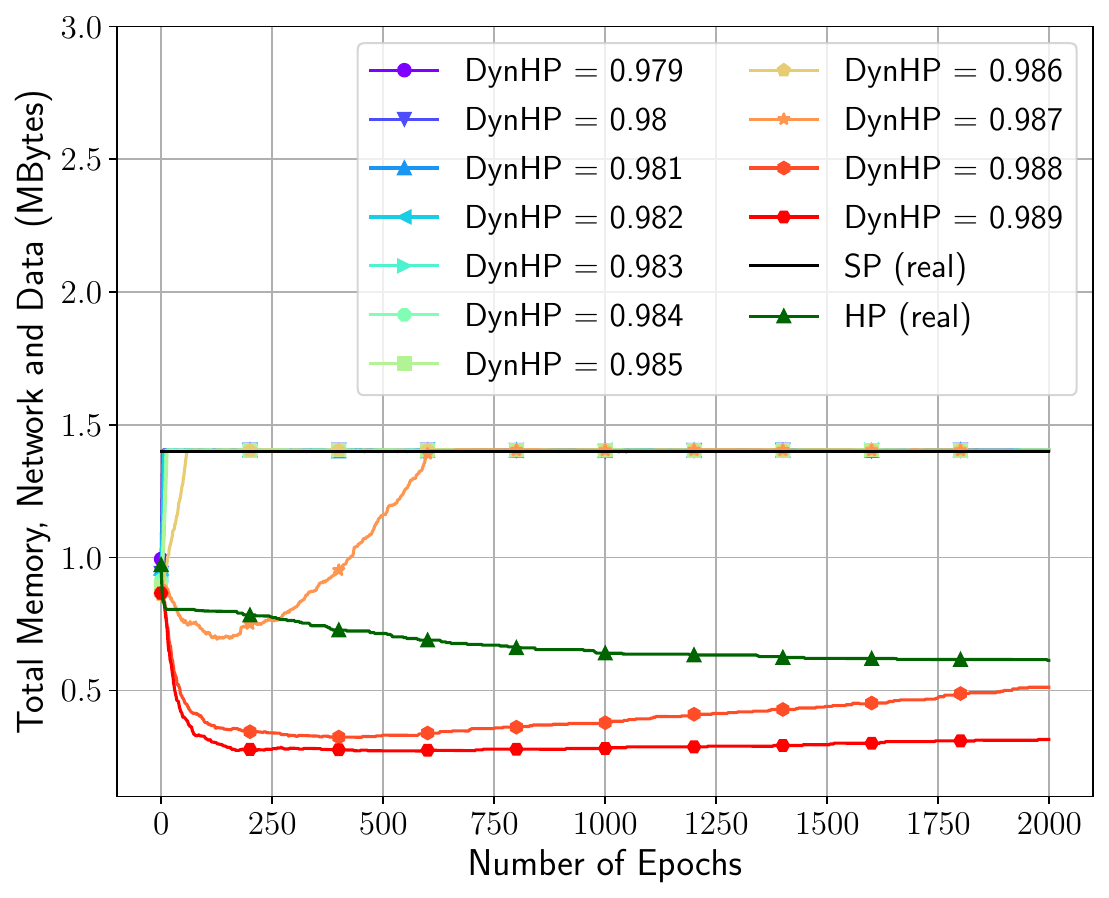}
	\caption{Fashion-MNIST\label{fig:fmnist_totmem}}
	\end{subfigure}
	~
	\begin{subfigure}[b]{0.48\textwidth}
	 \centering
	\includegraphics[width=1\textwidth]{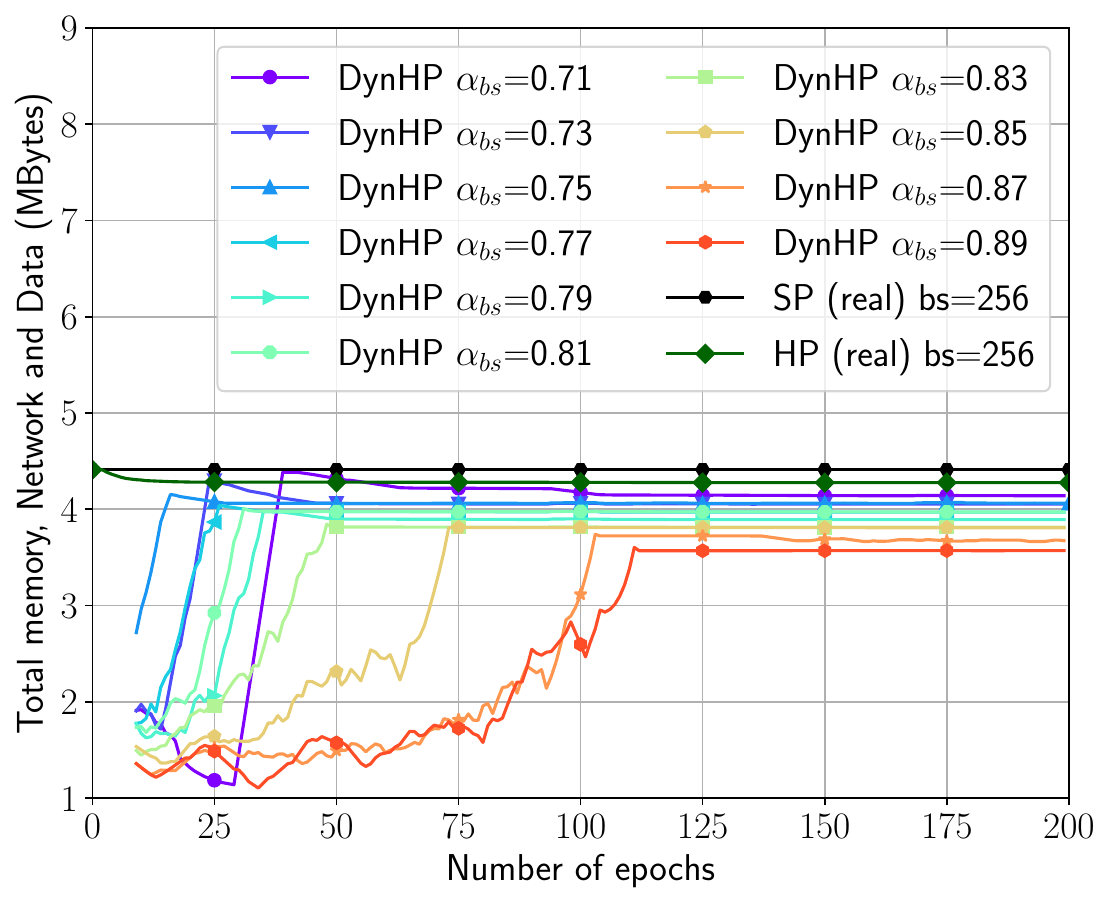}
	\caption{CIFAR-10\label{fig:cifar10_totmem}}
	\end{subfigure}
	\caption{Total memory occupation for DynHP with different configurations, HP (best) and SP (best).\label{fig:dynhp_totmem}}
\end{figure}

Finally, in Figure~\ref{fig:dynhp_totmem} we plot the total memory occupation for SP, HP, and DynHP that we measured during training epochs. For SP, we report the actual memory occupation, which is constant (solid black).
To clearly understand the hard-pruning behaviour without the influence of the dynamic batch sizing, we can observe the HP curve, where all the memory reduction is due to the hard-pruning procedure. As we can see, the pruning process is stepwise, i.e., after a pruning session, there is a plateau during which the network recovers the loss of expressive power, followed by another pruning session and another plateau.

Regarding DynHP, we see that, with proper values of $\alpha_{bs}$, it uses less memory than SP since higher values of $\alpha_{bs}$ limit the increment of the mini-batches. Moreover, we can appreciate how the dynamic sizing of the mini-batches combines with the HP procedure. Specifically, it is possible to distinguish when one of the two mechanisms dominates over the other one. In each curve's first trait, the mini-batches dynamic sizing is predominant over the hard pruning because the total memory occupation increases. Afterwards, we see that the HP mechanism causes a sheer drop of the total memory occupation, e.g., for MNIST around epoch 50 for DynHP with $\alpha_{bs}=0.972$ and for Fashion-MNIST at epoch 100 for $\alpha_{bs}=0.988$,  followed by a modest but still constant pruning. Note that the moment of the training at which this happens varies depending on the specific settings.
For what regards ResNet on CIFAR-10 (Figure \ref{fig:cifar10_totmem}), we observe a slightly different behaviour. We see that the parameter $\alpha_{bs}$ affects the growth rate of the total memory. For low $\alpha_{bs}$ values, we have a sheer increment of the total memory in the first $50$ training epochs, while for higher values, the growing effect is slower. Differently from MNIST and Fashion-MNIST, here, the size of mini-batches dominates the total memory occupation. Nonetheless, the growing effect on the total memory induced by the mini-batch shows that the pruning introduced by DynHP can save significant memory along with the training.

\subsubsection{Computational effort}
We now provide an analysis of the computational effort required to perform inference on the final trained models (Tables \ref{tab:flops} and \ref{tab:tot_flops}). Specifically, we compute for each model and dataset the expected flops, defined as the number of floating point operations needed to execute the model. The expected flops are computed as follows. For dense layers we have flops$_{dense}=(2*\mathrm{n}_{l}-1)*\mathrm{n}_{l+1}$, where $\mathrm{n}_{l},\mathrm{n}_{l+1}$is the number of (active) neurons of layer $l$ and $l+1$, respectively. For the convolutional layers $\mathrm{flops}_{conv}= \mathrm{fpf * nf}$, where $\mathrm{nf}$ represents the number of convolutional filters in the layer and $\mathrm{fpf}$ the flops count for one filter.

In Table \ref{tab:flops}, for each dataset, we report the flop count required to perform the inference on the final trained model. In this analysis, we consider only the best configurations of the three models, i.e., the ones with maximum accuracy.

Regarding DynHP, we selected $\alpha_{bs}=0.73,0.98,0.79$ for MNIST, Fashion-MNIST and CIFAR-10, respectively. As we can see, the three approaches succeed in learning models that actively reduce the flops count required to use them. In particular, while for MNIST, the final number of flops required by DynHP is close to the half ($46$\%) of the ones required by the original -- not pruned -- model, on Fashion-MNIST, DynHP allows to reduce the number of flops of $12$\% w.r.t. the original model. On CIFAR-10, the result is even more significant as DynHP significantly outperforms its counterparts by reducing the flops of up to 36\% w.r.t. the original model.

\begin{table}[h!]
	\centering
	\caption{Expected flops count for each dataset and model considering the best performing configuration. In (\%) we report the percentage of flops saved w.r.t. the solution without pruning.\label{tab:flops}}
	\small
	\begin{tabular}{lllll}
		\toprule
		& No pruning & SP & HP & DynHP \\
		& $(\cdot10^5)$ & $(\cdot10^5)$ & $(\cdot10^5)$ & $(\cdot10^5)$ \\	
		\midrule
		MNIST & $5.32$ $(-)$ & $3.06$ $(43\%)$ & $3.01$ $(44\%)$ & $2.87$ $(46\%)$ \\
		F-MNIST & $5.32$ $(-)$ & $4.01$ $(25\%)$ & $4.63$ $(13\%)$ & $4.67$ $(12\%)$ \\
		C-10 & $3.51\cdot 10^3$ $(-)$ & $3.38\cdot 10^3$ $(4\%)$ & $3.07\cdot 10^3$ $(13\%)$ & $2.25\cdot 10^3$ $(36\%)$ \\
		\bottomrule
	\end{tabular}
\end{table}

\begin{table}[h!]
	\centering
	\caption{Expected total flops count (sum over the epochs) for each dataset and model considering the best performing configuration. In (\%) we report the percentage of flops saved w.r.t. the solution without pruning.\label{tab:tot_flops}}
	\small
	\begin{tabular}{lllll}
		\toprule
		& No pruning & SP & HP & DynHP \\
		& $(\cdot10^7)$ & $(\cdot10^7)$ & $(\cdot10^7)$ & $(\cdot10^7)$ \\	
		\midrule
		MNIST & $1.06$ $(-)$ & $9.99$ $(6\%)$ & $7.84$ $(26\%)$ & $7.09$ $(33\%)$ \\
		F-MNIST & $1.06$ $(-)$ & $9.99$ $(6\%)$ & $9.74$ $(8\%)$ & $9.84$ $(8\%)$ \\
		C-10 & $7.03\cdot 10^3$ $(-)$ & $6.44\cdot 10^3$ $(8\%)$ & $6.16\cdot 10^3$ $(12\%)$ & $4.51\cdot 10^3$ $(36\%)$ \\
		\bottomrule
	\end{tabular}
\end{table}

Table \ref{tab:tot_flops} reports the sum of the flop counts over all the training epochs. Here, the benefits of employing DynHP at training time is apparent over all the datasets, where our method is able to achieve a reduction of the total flop count of more than one third ($36$\%) with respect to the flops needed to train the original -- not pruned -- model. These results confirm the effectiveness of DynHP in learning effective models while incrementally reducing not only the memory occupation but also the overall computational effort required for learning them.


\section{Conclusion}
\label{sec:conclusion}
We investigated the problem of learning compressed NNs with a fixed - and potentially small - memory budget on edge/fog devices. We proposed DynHP, a new resource-efficient NN learning technique that achieves performances comparable to conventional neural network training algorithms while enabling significant levels of network compression. DynHP prunes incrementally but permanently the network as the training process progresses by identifying neurons that contribute only marginally to the model's accuracy.
The memory saved during training is effectively reused by a \emph{dynamic batch sizing} technique that minimizes the accuracy degradation caused by the hard pruning strategy by adaptively adjusting the size of the data provided to the neural network so to improve its convergence and final effectiveness. The careful combination in DynHP of these two ingredients -- hard pruning and dynamic batch sizing -- allows us to train high-quality NN models by respecting the memory footprint constraints introduced by using resource-constrained edge/fog devices.  The extensive experiments conducted on two different kinds of neural network architectures, i.e., MLP and ResNet, on three public datasets, i.e., MNIST, FashionMNIST, and CIFAR-10, show that DynHP is able to compress the target neural network from $2$ to $10$ times without significant performance drops -- up to $3.5\%$ of additional misclassification error w.r.t. state-of-the-art competitors -- by reducing up to $80\%$ the overall memory occupation during training.

\medskip
\noindent \textbf{Future Directions}.
This work is a first step toward effective -- yet efficient and pervasive -- AI. In the future, we first intend to perform a formal convergence analysis that takes into account the joint combination of the two dynamics that characterize our proposed solution. Furthermore, we intend to investigate the idea of simultaneous training and compression of neural networks in a scenario involving distributed training at the edge of the network. We also want to investigate the application of quantization techniques in our solution and how it affects the efficiency-effectiveness trade-off of the models learned.

\section*{CRediT author statement}
\textbf{Lorenzo Valerio} and \textbf{Franco Maria Nardini}: Conceptualization, Methodology, Software, Validation, Writing - Original Draft. \textbf{Andrea Passarella} and \textbf{Raffaele Perego}: Conceptualization, Writing - Review \& Editing.

\section*{Acknowledgements}This work is partially supported by four projects: HumanE AI Network (EU H2020 HumanAI-Net, GA \#952026) Big Data to Enable Global Disruption of the Grapevine-powered Industries (EU H2020 BigDataGrapes, GA \# 780751), Operational Knowledge from Insights and Analytics on Industrial Data (MIUR PON OK-INSAID, ARS01\_00917), H2020 MARVEL (GA \#957337).

\section*{References}
\bibliographystyle{elsarticle-harv.bst}
\bibliography{./biblio.bib}

\begin{thebibliography}{36}
\expandafter\ifx\csname natexlab\endcsname\relax\def\natexlab#1{#1}\fi
\providecommand{\url}[1]{\texttt{#1}}
\providecommand{\href}[2]{#2}
\providecommand{\path}[1]{#1}
\providecommand{\DOIprefix}{doi:}
\providecommand{\ArXivprefix}{arXiv:}
\providecommand{\URLprefix}{URL: }
\providecommand{\Pubmedprefix}{pmid:}
\providecommand{\doi}[1]{\href{http://dx.doi.org/#1}{\path{#1}}}
\providecommand{\Pubmed}[1]{\href{pmid:#1}{\path{#1}}}
\providecommand{\bibinfo}[2]{#2}
\ifx\xfnm\relax \def\xfnm[#1]{\unskip,\space#1}\fi
\bibitem[{Ba and Caruana(2013)}]{Ba2013}
\bibinfo{author}{Ba, L.J.}, \bibinfo{author}{Caruana, R.},
  \bibinfo{year}{2013}.
\newblock \bibinfo{title}{{Do Deep Nets Really Need to be Deep?}} ,
  \bibinfo{pages}{1--9}\URLprefix \url{http://arxiv.org/abs/1312.6184},
  \href{http://arxiv.org/abs/1312.6184}{{\tt arXiv:1312.6184}}.
\bibitem[{Balles et~al.(2017)Balles, Romero and Hennig}]{Balles2017}
\bibinfo{author}{Balles, L.}, \bibinfo{author}{Romero, J.},
  \bibinfo{author}{Hennig, P.}, \bibinfo{year}{2017}.
\newblock \bibinfo{title}{Coupling adaptive batch sizes with learning rates},
  in: \bibinfo{booktitle}{Proceedings of the Thirty-Third Conference on
  Uncertainty in Artificial Intelligence, {UAI} 2017, Sydney, Australia, August
  11-15, 2017}.
\bibitem[{Barbalace et~al.(2020)Barbalace, Karaoui, Wang, Xing, Olivier and
  Ravindran}]{Barbalace:2020}
\bibinfo{author}{Barbalace, A.}, \bibinfo{author}{Karaoui, M.L.},
  \bibinfo{author}{Wang, W.}, \bibinfo{author}{Xing, T.},
  \bibinfo{author}{Olivier, P.}, \bibinfo{author}{Ravindran, B.},
  \bibinfo{year}{2020}.
\newblock \bibinfo{title}{Edge computing: The case for heterogeneous-isa
  container migration}, in: \bibinfo{booktitle}{Proceedings of the 16th ACM
  SIGPLAN/SIGOPS International Conference on Virtual Execution Environments},
  \bibinfo{publisher}{Association for Computing Machinery},
  \bibinfo{address}{New York, NY, USA}. pp. \bibinfo{pages}{73--87}.
\newblock \URLprefix \url{https://doi.org/10.1145/3381052.3381321},
  \DOIprefix\doi{10.1145/3381052.3381321}.
\bibitem[{Bellec et~al.(2017)Bellec, Kappel, Maass and Legenstein}]{Bellec2017}
\bibinfo{author}{Bellec, G.}, \bibinfo{author}{Kappel, D.},
  \bibinfo{author}{Maass, W.}, \bibinfo{author}{Legenstein, R.},
  \bibinfo{year}{2017}.
\newblock \bibinfo{title}{{Deep Rewiring: Training very sparse deep networks}},
  pp. \bibinfo{pages}{1--24}.
\newblock \URLprefix \url{http://arxiv.org/abs/1711.05136},
  \href{http://arxiv.org/abs/1711.05136}{{\tt arXiv:1711.05136}}.
\bibitem[{Conti et~al.(2017)Conti, Passarella and Das}]{Conti:2017aa}
\bibinfo{author}{Conti, M.}, \bibinfo{author}{Passarella, A.},
  \bibinfo{author}{Das, S.K.}, \bibinfo{year}{2017}.
\newblock \bibinfo{title}{The internet of people (iop): A new wave in pervasive
  mobile computing}.
\newblock \bibinfo{journal}{Pervasive and Mobile Computing}
  \bibinfo{volume}{41}, \bibinfo{pages}{1 -- 27}.
\newblock \URLprefix
  \url{http://www.sciencedirect.com/science/article/pii/S1574119217303723},
  \DOIprefix\doi{https://doi.org/10.1016/j.pmcj.2017.07.009}.
\bibitem[{Courbariaux et~al.(2015)Courbariaux, Bengio and
  David}]{Courbariaux2015}
\bibinfo{author}{Courbariaux, M.}, \bibinfo{author}{Bengio, Y.},
  \bibinfo{author}{David, J.P.}, \bibinfo{year}{2015}.
\newblock \bibinfo{title}{{BinaryConnect: Training Deep Neural Networks with
  binary weights during propagations}} , \bibinfo{pages}{1--9}\URLprefix
  \url{http://arxiv.org/abs/1511.00363}, \DOIprefix\doi{arXiv: 1412.7024},
  \href{http://arxiv.org/abs/1511.00363}{{\tt arXiv:1511.00363}}.
\bibitem[{Elsken et~al.(2019)Elsken, Metzen and Hutter}]{JMLR:v20:18-598}
\bibinfo{author}{Elsken, T.}, \bibinfo{author}{Metzen, J.H.},
  \bibinfo{author}{Hutter, F.}, \bibinfo{year}{2019}.
\newblock \bibinfo{title}{Neural architecture search: A survey}.
\newblock \bibinfo{journal}{Journal of Machine Learning Research}
  \bibinfo{volume}{20}, \bibinfo{pages}{1--21}.
\newblock \URLprefix \url{http://jmlr.org/papers/v20/18-598.html}.
\bibitem[{Frankle and Carbin(2019)}]{Frankle2018}
\bibinfo{author}{Frankle, J.}, \bibinfo{author}{Carbin, M.},
  \bibinfo{year}{2019}.
\newblock \bibinfo{title}{{The Lottery Ticket Hypothesis: Finding Sparse,
  Trainable Neural Networks}}.
\newblock \bibinfo{journal}{Proc. ICLR} , \bibinfo{pages}{1--42}\URLprefix
  \url{http://arxiv.org/abs/1803.03635},
  \href{http://arxiv.org/abs/1803.03635}{{\tt arXiv:1803.03635}}.
\bibitem[{Gal and Ghahramani(2016)}]{Gal2015}
\bibinfo{author}{Gal, Y.}, \bibinfo{author}{Ghahramani, Z.},
  \bibinfo{year}{2016}.
\newblock \bibinfo{title}{{Dropout as a Bayesian Approximation: Representing
  Model Uncertainty in Deep Learning}}, in: \bibinfo{booktitle}{Proc. 33rd Int.
  Conf. Int. Conf. Mach. Learn.}, pp. \bibinfo{pages}{1050--1059}.
\newblock \URLprefix \url{http://arxiv.org/abs/1506.02142},
  \href{http://arxiv.org/abs/1506.02142}{{\tt arXiv:1506.02142}}.
\bibitem[{Garcia~Lopez et~al.(2015)Garcia~Lopez, Montresor, Epema, Datta,
  Higashino, Iamnitchi, Barcellos, Felber and Riviere}]{Garcia-Lopez:2015aa}
\bibinfo{author}{Garcia~Lopez, P.}, \bibinfo{author}{Montresor, A.},
  \bibinfo{author}{Epema, D.}, \bibinfo{author}{Datta, A.},
  \bibinfo{author}{Higashino, T.}, \bibinfo{author}{Iamnitchi, A.},
  \bibinfo{author}{Barcellos, M.}, \bibinfo{author}{Felber, P.},
  \bibinfo{author}{Riviere, E.}, \bibinfo{year}{2015}.
\newblock \bibinfo{title}{Edge-centric computing: Vision and challenges}.
\newblock \bibinfo{journal}{SIGCOMM Comput. Commun. Rev.} \bibinfo{volume}{45},
  \bibinfo{pages}{37--42}.
\newblock \URLprefix \url{https://doi.org/10.1145/2831347.2831354},
  \DOIprefix\doi{10.1145/2831347.2831354}.
\bibitem[{Guo et~al.(2016)Guo, Yao and Chen}]{Guo2016b}
\bibinfo{author}{Guo, Y.}, \bibinfo{author}{Yao, A.}, \bibinfo{author}{Chen,
  Y.}, \bibinfo{year}{2016}.
\newblock \bibinfo{title}{{Dynamic Network Surgery for Efficient DNNs}}, in:
  \bibinfo{booktitle}{Intl. Conf. Neural Inf. Process.}
\newblock \URLprefix \url{http://arxiv.org/abs/1608.04493},
  \href{http://arxiv.org/abs/1608.04493}{{\tt arXiv:1608.04493}}.
\bibitem[{Han et~al.(2015)Han, Mao and Dally}]{Han2015}
\bibinfo{author}{Han, S.}, \bibinfo{author}{Mao, H.}, \bibinfo{author}{Dally,
  W.J.}, \bibinfo{year}{2015}.
\newblock \bibinfo{title}{{Deep Compression: Compressing Deep Neural Networks
  with Pruning, Trained Quantization and Huffman Coding}} ,
  \bibinfo{pages}{1--14}\URLprefix \url{http://arxiv.org/abs/1510.00149},
  \DOIprefix\doi{abs/1510.00149/1510.00149},
  \href{http://arxiv.org/abs/1510.00149}{{\tt arXiv:1510.00149}}.
\bibitem[{Han et~al.(2017)Han, Pool, Narang, Mao, Gong, Tang, Elsen, Vajda,
  Paluri, Tran, Catanzaro and Dally}]{Han2016a}
\bibinfo{author}{Han, S.}, \bibinfo{author}{Pool, J.}, \bibinfo{author}{Narang,
  S.}, \bibinfo{author}{Mao, H.}, \bibinfo{author}{Gong, E.},
  \bibinfo{author}{Tang, S.}, \bibinfo{author}{Elsen, E.},
  \bibinfo{author}{Vajda, P.}, \bibinfo{author}{Paluri, M.},
  \bibinfo{author}{Tran, J.}, \bibinfo{author}{Catanzaro, B.},
  \bibinfo{author}{Dally, W.J.}, \bibinfo{year}{2017}.
\newblock \bibinfo{title}{{DSD: Dense-Sparse-Dense Training for Deep Neural
  Networks}}, in: \bibinfo{booktitle}{Proc. ICLR}.
\newblock \URLprefix \url{http://arxiv.org/abs/1607.04381},
  \href{http://arxiv.org/abs/1607.04381}{{\tt arXiv:1607.04381}}.
\bibitem[{He et~al.(2016)He, Zhang, Ren and Sun}]{he2016deep}
\bibinfo{author}{He, K.}, \bibinfo{author}{Zhang, X.}, \bibinfo{author}{Ren,
  S.}, \bibinfo{author}{Sun, J.}, \bibinfo{year}{2016}.
\newblock \bibinfo{title}{Deep residual learning for image recognition}, in:
  \bibinfo{booktitle}{Proceedings of the IEEE conference on computer vision and
  pattern recognition}, pp. \bibinfo{pages}{770--778}.
\bibitem[{Hinton et~al.(2015)Hinton, Vinyals and Dean}]{Hinton2015}
\bibinfo{author}{Hinton, G.}, \bibinfo{author}{Vinyals, O.},
  \bibinfo{author}{Dean, J.}, \bibinfo{year}{2015}.
\newblock \bibinfo{title}{{Distilling the Knowledge in a Neural Network}} ,
  \bibinfo{pages}{1--9}\URLprefix \url{http://arxiv.org/abs/1503.02531},
  \href{http://arxiv.org/abs/1503.02531}{{\tt arXiv:1503.02531}}.
\bibitem[{Hubara et~al.(2016)Hubara, Soudry and Yaniv}]{Hubara2016}
\bibinfo{author}{Hubara, I.}, \bibinfo{author}{Soudry, D.},
  \bibinfo{author}{Yaniv, R.E.}, \bibinfo{year}{2016}.
\newblock \bibinfo{title}{{Binarized Neural Networks}} ,
  \bibinfo{pages}{1--9}\URLprefix \url{http://arxiv.org/abs/1602.02505},
  \href{http://arxiv.org/abs/1602.02505}{{\tt arXiv:1602.02505}}.
\bibitem[{Jin et~al.(2016)Jin, Yuan, Feng and Yan}]{Jin2016}
\bibinfo{author}{Jin, X.}, \bibinfo{author}{Yuan, X.}, \bibinfo{author}{Feng,
  J.}, \bibinfo{author}{Yan, S.}, \bibinfo{year}{2016}.
\newblock \bibinfo{title}{{Training Skinny Deep Neural Networks with Iterative
  Hard Thresholding Methods}} , \bibinfo{pages}{1--11}\URLprefix
  \url{http://arxiv.org/abs/1607.05423},
  \href{http://arxiv.org/abs/1607.05423}{{\tt arXiv:1607.05423}}.
\bibitem[{Kingma et~al.(2015)Kingma, Salimans and Welling}]{Kingma2015}
\bibinfo{author}{Kingma, D.P.}, \bibinfo{author}{Salimans, T.},
  \bibinfo{author}{Welling, M.}, \bibinfo{year}{2015}.
\newblock \bibinfo{title}{Variational dropout and the local reparameterization
  trick}, in: \bibinfo{booktitle}{Nips '15}, pp. \bibinfo{pages}{1--9}.
\newblock \href{http://arxiv.org/abs/1506.02557}{{\tt arXiv:1506.02557}}.
\bibitem[{Krizhevsky et~al.(2009)Krizhevsky, Hinton
  et~al.}]{krizhevsky2009learning}
\bibinfo{author}{Krizhevsky, A.}, \bibinfo{author}{Hinton, G.}, et~al.,
  \bibinfo{year}{2009}.
\newblock \bibinfo{title}{Learning multiple layers of features from tiny
  images} .
\bibitem[{Krizhevsky et~al.(2012)Krizhevsky, Sutskever and
  Hinton}]{krizhevsky2012imagenet}
\bibinfo{author}{Krizhevsky, A.}, \bibinfo{author}{Sutskever, I.},
  \bibinfo{author}{Hinton, G.E.}, \bibinfo{year}{2012}.
\newblock \bibinfo{title}{Imagenet classification with deep convolutional
  neural networks}, in: \bibinfo{booktitle}{Advances in neural information
  processing systems}, pp. \bibinfo{pages}{1097--1105}.
\bibitem[{LeCun(1998)}]{lecun1998mnist}
\bibinfo{author}{LeCun, Y.}, \bibinfo{year}{1998}.
\newblock \bibinfo{title}{The mnist database of handwritten digits}.
\newblock \bibinfo{journal}{http://yann. lecun. com/exdb/mnist/} .
\bibitem[{Lin et~al.(2016)Lin, Talathi and Annapureddy}]{Lin2016}
\bibinfo{author}{Lin, D.D.}, \bibinfo{author}{Talathi, S.S.},
  \bibinfo{author}{Annapureddy, V.S.}, \bibinfo{year}{2016}.
\newblock \bibinfo{title}{{F IXED P OINT Q UANTIZATION OF D EEP C ONVOLU -}},
  in: \bibinfo{booktitle}{ICLR}.
\bibitem[{Lin et~al.(2020)Lin, Stich, Barba, Dmitriev and
  Jaggi}]{DBLP:conf/iclr/LinSBDJ20}
\bibinfo{author}{Lin, T.}, \bibinfo{author}{Stich, S.U.},
  \bibinfo{author}{Barba, L.}, \bibinfo{author}{Dmitriev, D.},
  \bibinfo{author}{Jaggi, M.}, \bibinfo{year}{2020}.
\newblock \bibinfo{title}{Dynamic model pruning with feedback}, in:
  \bibinfo{booktitle}{{ICLR}}, \bibinfo{publisher}{OpenReview.net}.
\bibitem[{Louizos et~al.(2017)Louizos, Welling and Kingma}]{Louizos2017}
\bibinfo{author}{Louizos, C.}, \bibinfo{author}{Welling, M.},
  \bibinfo{author}{Kingma, D.P.}, \bibinfo{year}{2017}.
\newblock \bibinfo{title}{{Learning Sparse Neural Networks through
  {\$}L{\_}0{\$} Regularization}} , \bibinfo{pages}{1--13}\URLprefix
  \url{http://arxiv.org/abs/1712.01312},
  \href{http://arxiv.org/abs/1712.01312}{{\tt arXiv:1712.01312}}.
\bibitem[{Maddison et~al.(2017)Maddison, Mnih, Teh, Kingdom and
  Kingdom}]{Maddison2017}
\bibinfo{author}{Maddison, C.J.}, \bibinfo{author}{Mnih, A.},
  \bibinfo{author}{Teh, Y.W.}, \bibinfo{author}{Kingdom, U.},
  \bibinfo{author}{Kingdom, U.}, \bibinfo{year}{2017}.
\newblock \bibinfo{title}{{THe Concrete Distribution : a Continuous Relaxation
  of Discrete random variables}}, in: \bibinfo{booktitle}{ICLR}, pp.
  \bibinfo{pages}{1--17}.
\bibitem[{Modarressi and Sarbazi-Azad(2018)}]{Modarressi:2018aa}
\bibinfo{author}{Modarressi, M.}, \bibinfo{author}{Sarbazi-Azad, H.},
  \bibinfo{year}{2018}.
\newblock \bibinfo{title}{Chapter six - topology specialization for
  networks-on-chip in the dark silicon era}, in: \bibinfo{editor}{Hurson,
  A.R.}, \bibinfo{editor}{Sarbazi-Azad, H.} (Eds.), \bibinfo{booktitle}{Dark
  Silicon and Future On-chip Systems}. \bibinfo{publisher}{Elsevier}. volume
  \bibinfo{volume}{110} of \textit{\bibinfo{series}{Advances in Computers}},
  pp. \bibinfo{pages}{217 -- 258}.
\newblock \URLprefix
  \url{http://www.sciencedirect.com/science/article/pii/S0065245818300226},
  \DOIprefix\doi{https://doi.org/10.1016/bs.adcom.2018.03.009}.
\bibitem[{Molchanov et~al.(2017a)Molchanov, Ashukha and
  Vetrov}]{molchanov2017variational}
\bibinfo{author}{Molchanov, D.}, \bibinfo{author}{Ashukha, A.},
  \bibinfo{author}{Vetrov, D.}, \bibinfo{year}{2017}a.
\newblock \bibinfo{title}{Variational dropout sparsifies deep neural networks},
  in: \bibinfo{booktitle}{Proceedings of the 34th International Conference on
  Machine Learning-Volume 70}, \bibinfo{organization}{JMLR.org}. pp.
  \bibinfo{pages}{2498--2507}.
\bibitem[{Molchanov et~al.(2017b)Molchanov, Tyree, Karras, Aila and
  Kautz}]{Molchanov2016}
\bibinfo{author}{Molchanov, P.}, \bibinfo{author}{Tyree, S.},
  \bibinfo{author}{Karras, T.}, \bibinfo{author}{Aila, T.},
  \bibinfo{author}{Kautz, J.}, \bibinfo{year}{2017}b.
\newblock \bibinfo{title}{{Pruning Convolutional Neural Networks for Resource
  Efficient Inference}}.
\newblock \bibinfo{journal}{Proc. ICLR} , \bibinfo{pages}{1--17}\URLprefix
  \url{http://arxiv.org/abs/1611.06440},
  \href{http://arxiv.org/abs/1611.06440}{{\tt arXiv:1611.06440}}.
\bibitem[{Narang et~al.(2017)Narang, Elsen, Diamos and Sengupta}]{Narang2017}
\bibinfo{author}{Narang, S.}, \bibinfo{author}{Elsen, E.},
  \bibinfo{author}{Diamos, G.}, \bibinfo{author}{Sengupta, S.},
  \bibinfo{year}{2017}.
\newblock \bibinfo{title}{{Exploring Sparsity in Recurrent Neural Networks}},
  in: \bibinfo{booktitle}{Proc. ICLR}, pp. \bibinfo{pages}{1--10}.
\newblock \URLprefix \url{http://arxiv.org/abs/1704.05119},
  \href{http://arxiv.org/abs/1704.05119}{{\tt arXiv:1704.05119}}.
\bibitem[{Neklyudov et~al.(2017)Neklyudov, Molchanov, Ashukha and
  Vetrov}]{Neklyudov2017}
\bibinfo{author}{Neklyudov, K.}, \bibinfo{author}{Molchanov, D.},
  \bibinfo{author}{Ashukha, A.}, \bibinfo{author}{Vetrov, D.},
  \bibinfo{year}{2017}.
\newblock \bibinfo{title}{{Structured Bayesian Pruning via Log-Normal
  Multiplicative Noise}}.
\newblock \bibinfo{journal}{Intl. Conf. Neural Inf. Process.} ,
  \bibinfo{pages}{1--10}\URLprefix \url{http://arxiv.org/abs/1705.07283},
  \href{http://arxiv.org/abs/1705.07283}{{\tt arXiv:1705.07283}}.
\bibitem[{Smith et~al.(2018)Smith, Kindermans and Le}]{Smith2018}
\bibinfo{author}{Smith, S.L.}, \bibinfo{author}{Kindermans, P.J.},
  \bibinfo{author}{Le, Q.V.}, \bibinfo{year}{2018}.
\newblock \bibinfo{title}{Don't decay the learning rate, increase the batch
  size}, in: \bibinfo{booktitle}{International Conference on Learning
  Representations}.
\newblock \URLprefix \url{https://openreview.net/forum?id=B1Yy1BxCZ}.
\bibitem[{Srinivas and Babu(2015)}]{Srinivas2015}
\bibinfo{author}{Srinivas, S.}, \bibinfo{author}{Babu, R.V.},
  \bibinfo{year}{2015}.
\newblock \bibinfo{title}{{Data-free parameter pruning for Deep Neural
  Networks}} , \bibinfo{pages}{1--12}\URLprefix
  \url{http://arxiv.org/abs/1507.06149},
  \href{http://arxiv.org/abs/1507.06149}{{\tt arXiv:1507.06149}}.
\bibitem[{Srinivas et~al.(2017)Srinivas, Subramanya and Babu}]{Srinivas2017}
\bibinfo{author}{Srinivas, S.}, \bibinfo{author}{Subramanya, A.},
  \bibinfo{author}{Babu, R.V.}, \bibinfo{year}{2017}.
\newblock \bibinfo{title}{{Training Sparse Neural Networks}}.
\newblock \bibinfo{journal}{IEEE Comput. Soc. Conf. Comput. Vis. Pattern
  Recognit. Work.} \bibinfo{volume}{2017-July}, \bibinfo{pages}{455--462}.
\newblock \DOIprefix\doi{10.1109/CVPRW.2017.61}.
\bibitem[{Ullrich et~al.(2017)Ullrich, Meeds and Welling}]{ullrich2017soft}
\bibinfo{author}{Ullrich, K.}, \bibinfo{author}{Meeds, E.},
  \bibinfo{author}{Welling, M.}, \bibinfo{year}{2017}.
\newblock \bibinfo{title}{Soft weight-sharing for neural network compression}.
\newblock \bibinfo{journal}{arXiv preprint arXiv:1702.04008} .
\bibitem[{Xiao et~al.(2017)Xiao, Rasul and Vollgraf}]{xiao2017/online}
\bibinfo{author}{Xiao, H.}, \bibinfo{author}{Rasul, K.},
  \bibinfo{author}{Vollgraf, R.}, \bibinfo{year}{2017}.
\newblock \bibinfo{title}{Fashion-mnist: a novel image dataset for benchmarking
  machine learning algorithms}.
\newblock \href{http://arxiv.org/abs/cs.LG/1708.07747}{{\tt
  arXiv:cs.LG/1708.07747}}.
\bibitem[{Zhang et~al.(2016)Zhang, Bengio, Hardt, Recht and
  Vinyals}]{zhang2016understanding}
\bibinfo{author}{Zhang, C.}, \bibinfo{author}{Bengio, S.},
  \bibinfo{author}{Hardt, M.}, \bibinfo{author}{Recht, B.},
  \bibinfo{author}{Vinyals, O.}, \bibinfo{year}{2016}.
\newblock \bibinfo{title}{Understanding deep learning requires rethinking
  generalization}.
\newblock \bibinfo{journal}{arXiv preprint arXiv:1611.03530} .

\end{thebibliography}
\end{document}